\documentclass[%
 reprint,
 amsmath,amssymb,
 aps,
 prx,
 superscriptaddress,
 showkeys
]{revtex4-2}

\usepackage{xcolor}
\usepackage{graphicx}
\usepackage{dcolumn}
\usepackage{bm}
\usepackage{hyperref}

\newcommand{\x}{\boldsymbol{x}}

\newcommand{\spin}{\boldsymbol{s}}
\newcommand{\basis}{\boldsymbol{b}}
\newcommand{\dataset}{\boldsymbol{\hat{s}}}
\newcommand{\vecmu}{\boldsymbol{\mu}}
\newcommand{\vecnu}{\boldsymbol{\nu}}
\newcommand{\vecghat}{\boldsymbol{\hat{g}}}
\newcommand{\vecg}{\boldsymbol{g}}
\newcommand{\M}{\mathcal{M}}
\newcommand{\FIM}{\boldsymbol{I}}
\newcommand{\Tmat}{\boldsymbol{T}}

\begin{document}

\preprint{APS/123-QED}

\title{Bayesian Inference of Minimally Complex Models\\ with Interactions of Arbitrary Order}

\author{Cl\'elia de Mulatier}
 \email{c.m.c.demulatier@uva.nl}
 \affiliation{University of Amsterdam, Institute for Theoretical Physics, and Informatics Institute, Science Park 904,
1098 XH Amsterdam, the Netherlands}
    \affiliation{University of Pennsylvania, Department of Physics and Astronomy, Philadelphia, PA 19104, United States}
\author{Matteo Marsili}%
\affiliation{%
 The Abdus Salam International Centre for Theoretical Physics (ICTP), Strada Costiera 11, I-34014 Trieste, Italy
}

\date{\today}

\begin{abstract}
Finding the model that best describes a high-dimensional dataset is a daunting task, even more so if one aims to consider all possible high-order patterns of the data (i.e., correlation patterns between three or more variables), going beyond pairwise models.
For binary data, we show that this task becomes feasible when restricting the search to a family of {\em simple} models, that we call Minimally Complex Models (MCMs). MCMs are maximum entropy models that have interactions of arbitrarily high order grouped into independent components of minimal complexity. They are simple in information-theoretic terms, which means they can only fit well certain types of data patterns and are therefore easy to falsify. We show that Bayesian model selection restricted to these models is computationally feasible and has many advantages. First, the model evidence, which balances goodness-of-fit against complexity, can be computed efficiently without any parameter fitting, enabling very fast explorations of the space of MCMs. 
Second, the family of MCMs is invariant under gauge transformations, 
which can be used to develop a representation-independent approach to statistical modeling.
For small systems (up to 15 variables), combining these two results
allows us to select the best MCM among all, even though the number of models is already extremely large 
($10^{68}$ models for $15$ variables).  
For larger systems, 
we propose simple heuristics to find optimal MCMs in reasonable times.
Besides, inference and sampling can be performed without any computational effort. 
Finally, because MCMs have interactions of any order, they can reveal the presence of important high-order dependencies in the data, 
providing a new approach to explore high-order dependencies in complex systems.
We apply our method to synthetic data and real-world examples, 
illustrating how MCMs portray the structure of dependencies among variables in a simple manner, extracting falsifiable predictions on symmetries and invariance from the data.
\end{abstract}


\maketitle


\section{\label{sec:level1}Introduction}

``All models are wrong, but some models are useful''~\cite{box1976science}. This statement is particularly appropriate in the context of statistical inference of high-dimensional data. Recent spectacular advances in machine learning have shown that very complex models, such as deep neural networks \cite{lecun2015deep}, can be very ``useful'' in learning hidden features in high-dimensional data, making it possible to generalize from examples. 
Models that encode the Laws of Nature refer to a different notion of ``usefulness". 
As argued by Wigner~\cite{wigner1990unreasonable}, they describe regularities -- such as how bodies fall under the effect of gravity -- in {\it simple} forms involving only a few variables and parameters. 
These regularities occur in ways that are independent of many conditions that could affect them. 
Simple models, such as Newton's law, tell us more about independence than about dependence.
Their simplicity reflects specific principles -- such as invariances, symmetries, and conservation laws -- that are easy to falsify.
In this paper, we are interested in performing statistical inference for binary data with models that display such simplicity.

\begin{figure}
    \centering
    \includegraphics[width=.9\linewidth]{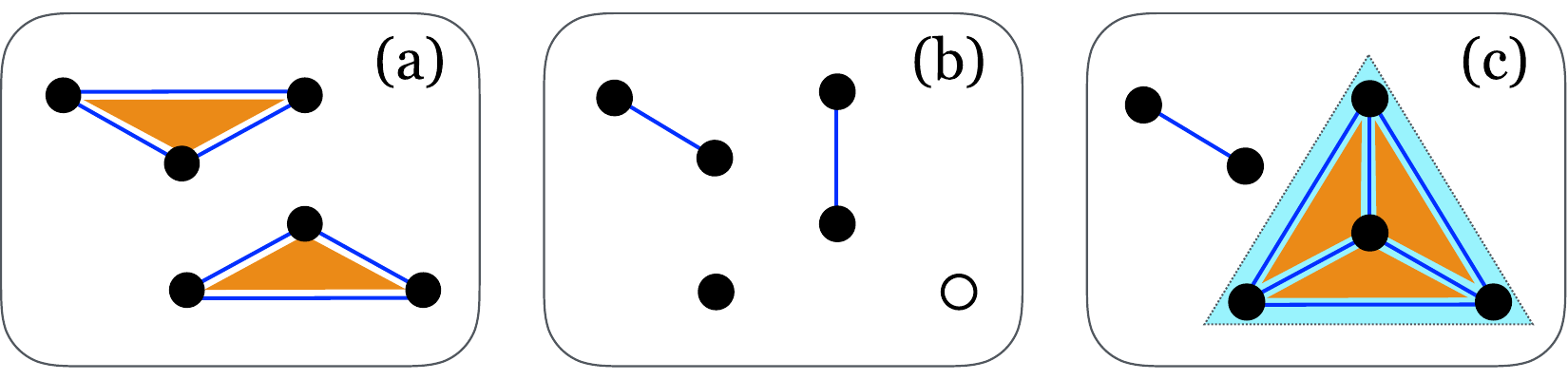}
\caption{(colors online)
    {\bf Examples of Minimally Complex Models} for a $6$-spin system. 
    Models are represented by diagrams: single-spin variables are dots, full in the presence of a local field and empty otherwise; pairwise interactions are blue lines; 3-spin interactions are orange triangles; and 4-spin interactions are light blue polygons enclosing four spins.
    Note that in this figure we show special examples of MCMs that correspond to partitions of the variables into independent parts, with the variables being completely connected within each part and not connected at all between the parts.
    Models of this type form the category MCM$^{*}$ in Fig.~\ref{Fig2:NbModels}.
    }
\label{Fig1:MCM_ex}
\end{figure}

In information theory, the {\it simplicity} of a statistical model is defined unambiguously in terms of the Minimum Description Length (MDL) complexity, which is the minimal number of bits required to describe the model~\cite{grunwald2007minimum}. 
This complexity is a measure of the number of different datasets that can be described by the same parametric model~\cite{myung2000counting}, and therefore, of the flexibility of the model.  
A {\it simple} model thus fits well few types of statistical structures of data
and, as a consequence, is easy to falsify. 
In Bayesian inference, the MDL complexity predicts how much a model should be penalized due to its flexibility~\cite{myung2000counting}. 
In the context of binary data, Beretta {\it et al.}~\cite{beretta2018stochastic} have studied the MDL complexity of spin models, which are maximum entropy models 
with interactions of arbitrary order.
They argue that, with the same number of parameters, the simplest models are those for which statistical dependencies concentrate on the smallest subset of variables. 
By contrast, spin models with pairwise interactions, such as the Ising model, turn out to be very complex. Their flexibility~\cite{merchan2016sufficiency} allows them to model a broad variety of datasets, from neuronal activity~\cite{schneidman2006weak, savin2017maximum} to voting outcomes~\cite{lee2015statistical}.
Yet, statistical inference of pairwise models is known to be computationally challenging \cite{montanari2009graphical, cocco2012adaptive, barton2016ace,vuffray2016interaction,nguyen2017inverse}, whereas the simplest spin models, called ``sub-complete models'' in~\cite{beretta2018stochastic},
are very easy to infer.\\
\indent In this paper, we show that the evidence of sub-complete models is also straightforward to compute and that this property makes Bayesian Model Selection (BMS) possible within a remarkably broad family of simple models for binary data, for which both the maximum likelihood and the model evidence are easy to compute. 
These models, which we call {\it Minimally Complex Models} (MCMs), 
are spin models formed of independent components of minimal complexity. 
In some representations, an MCM corresponds to a partition of the variables into independent parts (or ``components'')  
with no interactions between the parts and with all possible interactions within each part (see Fig.~\ref{Fig1:MCM_ex}). 
The best MCM for a given dataset thus captures the global structure of dependencies and independencies in the data,
providing robust predictions on dependencies between the variables.
In particular, our approach can identify the presence of important high-order interactions (i.e, between three or more variables).
Besides being simple and easy to infer, MCMs also enjoy properties that make statistical inference invariant under changes in the representation of data. Finally, inferred MCMs can also be sampled from with minimal computational effort.

In what follows, we formally define MCMs and discuss how to compare them in the context of BMS. As the set of potential MCMs to compare is huge (see Fig.~\ref{Fig2:NbModels}), we propose different heuristics 
to find the best MCM for a given dataset.
Our results are finally illustrated in several test cases, where one can appreciate how our approach uncovers global data structures of 
dependencies and independencies that are consistent with the discussed phenomena.

\section{Bayesian Model Selection of Spin Models}

A binary dataset $\dataset=(\spin^{(1)},\ldots,\spin^{(N)})$ is a set of $N$ independent observations of spin configurations $\spin=(s_1,\ldots,s_n)$, where $s_i=\pm 1$. In the following, we assume that each configuration $\spin^{(i)}$ is independently drawn from an unknown distribution, which we aim to infer. To do so, we consider the following parametric family of probability distributions:
\begin{equation}\label{eq:graph_model}
p(\spin\,|\,\vecg,\mathcal{M})
    =\frac{1}{Z_{\M}(\vecg)}\,\exp \left({\;\sum_{\vecmu\in \M} 
    g_{\vecmu}\phi^{\vecmu}(\spin)}\right)\,,
\end{equation}
where $\M=\{\vecmu^{(1)},\,\ldots, \vecmu^{(K)} \}$ is a set of $K$ interactions of arbitrary order, and where the $\vecmu^{(i)}$'s are binary vectors encoding which spins are involved in each interaction. 
Any choice of the set $\M$ of interactions defines a spin model. 
The operators 
$\phi^{\vecmu}(\spin)=\prod_{i=1}^{n} {s_i}^{\mu_i}\,$ are the product spin operators associated with the interactions $\vecmu=(\mu_1,\,\ldots,\,\mu_n)$, while the conjugate parameters $g_{\vecmu}$ modulate the strength of the interactions. 
For example, pairwise spin models contain operators of the form $\phi^{\vecmu}(\spin)=s_is_j$ between pairs of spins. 
In a 4-spin system, a three-body interaction between $s_1$, $s_2$, and $s_4$ is encoded by the binary vector $\vecmu=(1,1,0,1)$ and is associated with the operator $\phi^{\vecmu}(\spin) = s_1s_2s_4$.
Finally, the partition function $Z_\mathcal{M}(\vecg)$ ensures normalization (see App.~\ref{app:spinModels} for more details on spin models).

\begin{figure}
    \centering
    \includegraphics[width=\linewidth]{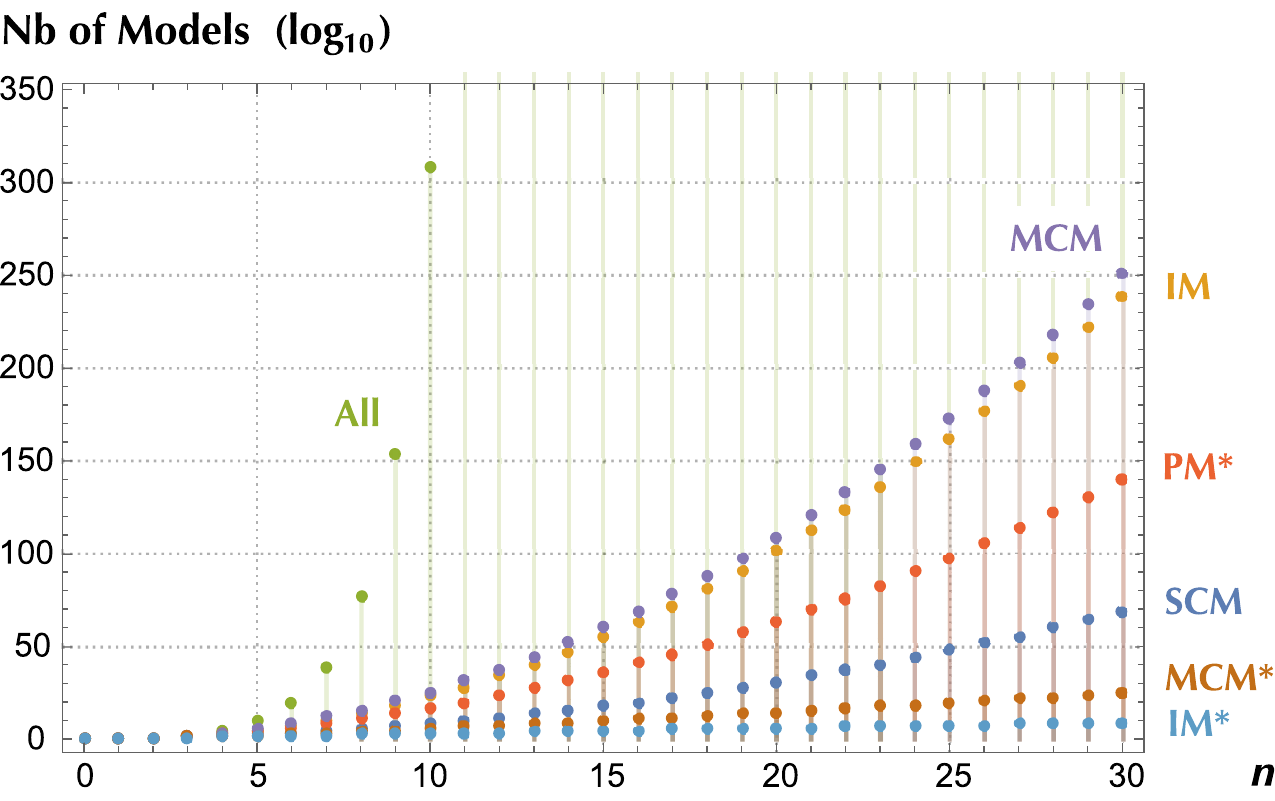}
\caption{(colors online)
    {\bf Number of spin models} as a function of the system size~$n$ for different families (see App.~\ref{app:enum} for proofs of the counts):
    all spin models (green), all Minimally Complex Models (MCM, violet), 
    all Independent Models (IM, orange),  
    and all Sub-Complete Models (SCM, dark blue).
    MCM$^*$ and IM$^*$ indicate respectively the number of MCM and IM that share the same preferred basis.
    For comparison, we also report the number of Pairwise Models (PM$^*$). 
    The number of IM and of MCM grows exponentially with $n$, roughly as $2^{n^2}$, whereas the number of PM$^*$ grows as $2^{n^{2}/2}$.
    The number of MCM$^*$ grows slower than $n^n$.
    Note that the $y$-axis reports the logarithm base 10 of the number of models.
    For example, at $n=9$, there are of the order of $10^{153}$ models, but only $10^{20}$ MCM, which include $10^{18}$ IM and $10^5$ MCM$^*$; there are $10^{13}$ PM$^*$.
    }
\label{Fig2:NbModels}
\end{figure}

Equation~(\ref{eq:graph_model}) defines a complete family of models, capable of describing all possible patterns of binary data with an appropriate choice of the set $\M$ of operators~\cite{mastromatteo2013typical}.
In absence of any prior knowledge about the system, one must compare the performance of all spin models to find which best describes a dataset~$\dataset$.
We focus on the best model~$\M$, which is the one achieving the largest posterior probability $P(\M\,|\,\dataset)$. This is obtained with Bayes' theorem from the {\it evidence} (or {\em marginal likelihood}) defined as follows~\cite{kruschke2014doing}:
\begin{equation}\label{eq:evidence}
P(\dataset\,|\,\M)
    =
    \int_{\mathbb{R}^M}\!d\vecg \,\prod_{i=1}^N \,p\big(\spin^{(i)}|\,\vecg,\mathcal{M}\big)\;P_0(\vecg\,|\,\M)\,,
\end{equation}
where $P_0(\vecg\,|\,\mathcal{M})$ is a prior distribution over the parameters. 
We assume a uniform prior over the models $\M$, such that the best model for data is also the one that maximizes the evidence.
Furthermore, we assume that $P_0(\vecg\,|\,\M)$  takes the form of Jeffreys' prior~\cite{jeffreys1946invariant}. With this choice, it was shown~\cite{myung2000counting} that in the limit of large datasets ($N\to\infty$) the model $\M$ maximizing the evidence in Eq.~\eqref{eq:evidence} is also the one providing the most succinct description of the data as defined by the Minimum Description Length (MDL) principle~\cite{rissanen1986stochastic, rissanen1996fisher, grunwald2007minimum}. 
Indeed, expanding Eq.~\eqref{eq:evidence} to order $O(1)$ in $N$ gives~\cite{myung2000counting}:
\begin{equation}\label{eq:MDL}
    \hspace{-0.5mm}
	\log P(\dataset\,|\,\M) \underset{N\to\infty}{\simeq} \log P(\dataset\,|\,\vecghat,\M)
		-
		\frac{K}{2}\log \left(\frac{N}{2\pi}\right)-c_{\M},
\end{equation}
where $\vecghat$ is the vector of maximum likelihood parameters. 
Using this expansion, 
the best model is the one that has the largest maximum log-likelihood penalized by two terms, which can be recognized as the MDL complexity of model $\M$.
The first is proportional to the number $K$ of parameters in the model (the more parameters, the more complex), while the second term $c_{\M}$ 
penalizes models depending on their geometry (see definition in App.~\ref{App:Complexity}).

Despite the existence of clear theoretical prescriptions on how to select the best model,
it is in practice nearly impossible to find it.
Indeed, it is computationally challenging to compute the evidence in Eq.~\eqref{eq:evidence} or the complexity~$c_{\mathcal{M}}$ for a generic spin model $\mathcal{M}$.
In addition, the number of potential models to compare is huge: in an $n$-spin system, there exist $2^n-1$ interactions of arbitrary order, and therefore $2^{2^n-1}$ potential models.
An exhaustive search among all spin models is unfeasible even for small systems (see Fig.~\ref{Fig2:NbModels}). 
A possible solution to this hurdle is to restrict model selection to a smaller family of models, such as pairwise models. 
For this family, model selection turns into a problem of graph reconstruction~\cite{montanari2009graphical}, that aims at identifying the network of interactions among spins. 
A plethora of methods and results have been derived within this framework~\cite{cocco2012adaptive, barton2016ace,nguyen2017inverse,vuffray2016interaction, bulso2016sparse, tkavcik2014searching}. 
Here, we propose to focus on a different family of models, the {\it minimally complex models} (MCMs).

\begin{figure}[h!]
    \centering
    \includegraphics[width=.96\linewidth]{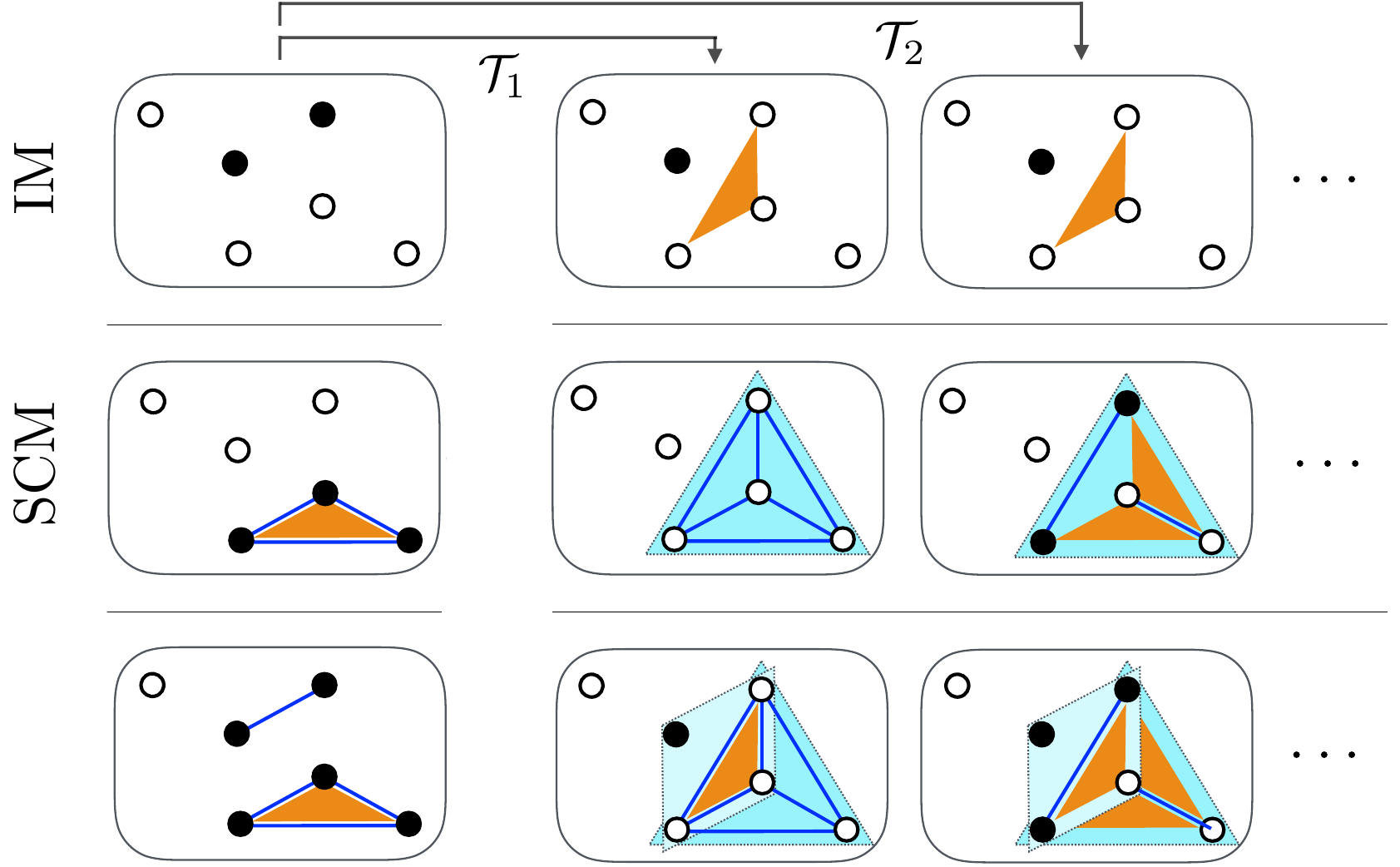}
\caption{
    (colors online)
    {\bf Examples of MCMs and basis transformations.}
    All these models are MCMs and are represented with the same notations as in Fig.~\ref{Fig1:MCM_ex}.
    In the first column, the first model is an independent model (IM) composed of two independent operators, the second is a sub-complete model (SCM) composed of a single independent complete component (ICC), and the last one is an MCM composed of two ICCs. 
    In each column,  models of each row are obtained by the same basis transformation~$\mathcal{T}_i$ of the models of the first column, 
    and therefore have the same properties as their respective original model (i.e., they are respectively IM, SCM, and MCM with two ICCs
    -- see Fig.~\ref{fig:SI:MCM_GT} in Appendix for details on the transformations $\mathcal{T}_i$ used).
    An alternative way of thinking about this is that
    each row displays the same abstract spin model represented in different bases. 
    }
\label{Fig3:MCM_ex_GT}
\end{figure}

\section{Minimally Complex Models (MCMs)}
\subsection{Definition}
MCMs are a subset of the spin models defined in Eq.~(\ref{eq:graph_model}), for which the set~$\M$ of interactions is the union of {\it independent complete components} (ICCs)~$\mathcal{M}_a$:
\begin{equation}
\label{eq:defMCM}
    \mathcal{M}=\bigcup_{a\in\mathcal{A}}\mathcal{M}_a\,,
\end{equation}
where $\mathcal{A}$ is the set of ICCs of $\M$.
A {\it complete component}~$\mathcal{M}_a$ is a set of interactions verifying that, for any two interactions $\vecmu,\vecnu\in\mathcal{M}_a$, the interaction $\vecmu\oplus\vecnu$ associated with the spin operator $\phi^{\vecmu\oplus\vecnu}(\spin)\doteq\phi^{\vecmu}(\spin)\,\phi^{\vecnu}(\spin)$ also belongs to~$\mathcal{M}_a$
($\oplus$ denotes the element-wise sum modulo 2 of the binary vectors $\vecmu$ and $\vecnu$ -- see App.~\ref{app:spinModels}).
Furthermore, complete components are said to be {\it independent} if none of the operators of a given component $\M_a$ can be obtained by combinations of operators of the other components (i.e. by products of operators of the set $\M\backslash\M_a$).
Figure~\ref{Fig3:MCM_ex_GT} shows several examples of MCMs, including two special cases: independent models and sub-complete models.
Independent models (IMs) are MCMs for which all ICCs have only a single operator. 
Sub-complete models (SCM) are MCMs with a single ICC, and they are the models with the smallest complexity in Ref.~\cite{beretta2018stochastic}.

\subsection{Preferred basis of an MCM}
Ref.~\cite{beretta2018stochastic} introduced the notion of equivalence classes of spin models. 
Models of the same class can be obtained from one another by certain basis transformations of the spin variables that preserve the structure of spin operators (see examples in Fig.~\ref{Fig3:MCM_ex_GT}) and have thus the same statistical properties~\footnote{Ref.~\cite{beretta2018stochastic} shows that models of the same class have the same Fisher information matrix up to permutations of their parameters.}. Such transformations are called {\em gauge transformation} in~\cite{beretta2018stochastic} (see App.~\ref{App:GT}).
From a different perspective, models of the same class can also be seen as the same abstract statistical model represented in different bases.
Each row in Fig.~\ref{Fig3:MCM_ex_GT} gives examples of models belonging to the same class. Here, one can observe that 
there are bases in which the representation of an MCM is easy to define 
in terms of a partition of the basis variables (see first column in Fig.~\ref{Fig3:MCM_ex_GT}).
In the following, we define this idea more formally, by introducing the notion of {\it preferred basis} of an MCM.

A {\it basis} $\basis$ of a model $\M$ is a minimal set of $r$~independent operators, $\basis=(\phi_1,\cdots,\phi_r)$, that can generate all the operators of $\M$. 
Thus, any operators $\phi\in\M$ can be written as a product of elements of~$\basis$ in a unique way. The cardinality~$r$ of such basis is called the {\it rank} of $\M$ (necessarily $r\leq n$).
We recall that a set~$\basis$ of operators is said {\it independent} if none of its operators can be obtained as a product of other operators of~$\basis$~\cite{beretta2018stochastic} (such a set of operators forms an IM).
For example, $\basis=\{s_1, s_2, s_3\}$ and $\basis'=\{s_1, s_1 s_2, s_1 s_2 s_3\}$ are two sets of independent operators, and both sets are possible bases for the model $\M=\{s_1, s_2, s_1 s_2, s_3\}$.

By definition, a {\it complete} component $\M_a$ is a special case of spin models that contains all the $2^{r_a}-1$ operators generated by its basis $\basis_a=(\phi_1, \cdots, \phi_{r_a})$.
As a direct consequence, the number of operators of an MCM is
\begin{align}
    K=\sum_{a\in\mathcal{A}} (2^{r_a}-1)\,.
\end{align}
Note that the choice of basis of an ICC is not unique. 
In the example above, $\basis$ and $\basis'$ both generate the same ICC 
$\M_a = \{s_1,  s_2, s_3, s_1 s_2, s_2 s_3, s_1 s_3, s_1 s_2 s_3\}$.

Finally, as a result of the {\it independence} between the complete components $\M_a$ of an MCM $\M$, the union of the bases~$\basis_a$ for each $\M_a$ forms a basis~$\basis$ of $\M$,
\begin{align}\label{eq:MCM:basis}
    \basis=\cup_{a\in\mathcal{A}}\basis_a\,,
\end{align}
and the rank of an MCM is $r=\sum_{a\in\mathcal{A}}r_a$.
We refer to such basis as a {\it preferred basis} of an MCM, because the representation of $\M$ precisely corresponds to a partition of the basis variables into ICCs (such as in the examples of Fig.~\ref{Fig1:MCM_ex}). 
Note that the preferred basis of an MCM is not unique (except for IMs).

\subsection{Properties of MCMs}
Let us now discuss the properties of MCMs. 
We first observe that, as a consequence of~Eq.~\eqref{eq:MCM:basis}, 
the state probability distribution of an MCM~$\M$ factorizes over the probability distributions of its ICCs, when expressed in terms of the operators $\basis(\spin)=\cup_a\,\basis_a(\spin)$ of a preferred basis of $\M$~(see App.~\ref{App:Factorization}): 
\begin{equation}
\label{factorization}
p(\spin\,|\,\vecg,\mathcal{M})=\frac{1}{2^{(n-r)}}\,\prod_{a\in\mathcal{A}}
\,p_a\big(\basis_a(\spin)\,|\,\vecg_a^\prime,\mathcal{M}_a^\prime\big).
\end{equation}
The ICCs $\M_a^\prime$ are the gauge transformed of the ICCs $\M_a$ of $\M$ in the basis $\basis_a$ 
and each $p_a$ is a probability distribution over $r_a$ spin variables only.
The prefactor corresponds to a $1/2$ probability for each of the $(n-r)$ variables not modeled by~$\M$ 
(such as the spins represented by empty dots in the first column of Fig.~\ref{Fig3:MCM_ex_GT}).
Equation~(\ref{factorization}) implies that both the likelihood and the evidence of an MCM factorize over its ICCs.
Note that this doesn't mean that these quantities are 
factorizable in the variables $\spin$ of the original dataset.
In addition, the evidence of an ICC is remarkably simple to compute analytically (see App.~\ref{Materials:Evidence}), 
and leads to this expression for the evidence of an MCM:
\begin{equation}
\label{eq:evidenceMCM}
    P(\dataset\,|\,\mathcal{M})
    =\frac{1}{2^{N(n-r)}}\prod_{a\in\mathcal{A}}\left[\frac{\Gamma(2^{r_a-1})}{\Gamma(N+2^{r_a-1})}\prod_{\basis_a}\left[\frac{\Gamma(k_{\basis_a}+\frac{1}{2})}{\Gamma(\frac{1}{2})}\right]\right]
\end{equation}
in which $N$ is the number of datapoints in $\dataset$, $\Gamma$ is the gamma function, and $k_{\basis_a}$ is the number of times that the basis operators of $\M_a^{\prime}$ take the value ${\basis_a}$ over the dataset. 
The product over $\basis_a$ denotes a product over all the $2^{r_a}$ values that can be taken by $\basis_a$;
states not observed in the data ($k_{\basis_a}=0$) 
contribute a factor~$1$. 
Finally, the value of the parameters that maximizes the likelihood function 
is also straightforward to compute and doesn't require any parameter fitting
(see App.~\ref{Materials:Evidence}).
As a consequence, the model probability distribution at the maximum-likelihood estimate $\boldsymbol{\hat{g}}$ also takes a very simple form 
(see App.~\ref{Materials:Evidence}):
\begin{equation}
\label{eq:MCM_P_s:maxL}
    p(\spin\,|\,\boldsymbol{\hat{g}},\M)=\frac{1}{2^{(n-r)}} 
\prod_{a\in\mathcal{A}}\frac{k_{\basis_a(\spin)}}{N},
\end{equation}
where $\basis_a(\spin)$ is the representation of $\spin$ in the basis $\basis_a$ 
and $k_{\basis_a(\spin)}$ is the number of time the state $\basis_a(\spin)$ is observed in the data.
This makes sampling from the model distribution computationally efficient (see App.~\ref{Materials:Sampling}). 
In summary, all the useful quantities in the context of statistical inference of MCM (maximum-likelihood estimate, value of the maximum likelihood and of the evidence, and model distribution at best fit) have simple expressions that can be readily computed from the counts $k_{\basis_a}$ of the number of times the configuration ${\basis_a}$ of the basis operators is observed in the data.

The invariance of the internal structure of MCMs with respect to {\it gauge transformations} (GTs) that we alluded to above deserves further discussion. 
A GT $\mathcal{T}$ is a bijection from the set of states to itself (automorphism),
where a new state $\spin^{\prime}$ is defined as a function of the original state $\spin$ by a set of $n$ independent operators: $\spin'=\mathcal{T}(\spin)=\left(\phi_1(\spin),\ldots,\phi_n(\spin)\right)$. As discussed above, GTs are changes of the reference frame with which data, operators or models are described. Indeed, GTs are 
bijections from the set of all operators to itself~\footnote{A transformed operator $\phi'$ 
is obtained from the original operator $\phi$ 
by $\phi'(\spin') = \phi(\mathcal{T}^{-1}(\spin'))$.}, and therefore also from the set of all spin models to itself~\footnote{A transformed model $\M'=\mathcal{T}(\M)$ is obtained by transforming each operator of $\M$.
} (see App.~\ref{App:GT}). The order of an operator, i.e. the number of spins that occur in it, is not invariant under GTs (as illustrated in Fig.~\ref{Fig3:MCM_ex_GT}).
However, the mutual relation between the operators of a model~\footnote{This mutual relation between operators of a model is called the loop structure of the model in Ref.~\cite{beretta2018stochastic}. 
Models with the same loop structure have the same Fisher information matrix.
} is preserved under a GT~\cite{beretta2018stochastic}, 
i.e., if $\phi_1(\spin)$ and $\phi_2(\spin)$ are two operators and $\phi_{1+2}(\spin)=\phi_1(\spin)\,\phi_2(\spin)$, then the gauge transformed operator $\phi_{1+2}'(\spin')$ is still the product of the transformed operators $\phi_1'(\spin')$ and $\phi_2'(\spin')$.
Therefore the invariance of the family of MCMs under GTs makes the space of MCMs
akin to a {\em cognitive map}, as defined e.g. in Ref.~\cite{whittington2022build}, endowed with an inbuilt relational structure that is an essential element in learning. 
It allows one to learn the relational structure between operators -- the sufficient statistics in models of the form of Eq.~\eqref{eq:graph_model} -- irrespective of how the data is represented.
This is not true for pairwise models, because the family of pairwise models is not closed under GTs: a GT applied to a pairwise model does not necessarily return a pairwise model.
Invariance under GTs of the considered family of models is a key requirement in BMS, because it makes it possible to uncover a hidden relational structure that is independent of how the data is represented.

Finally, Ref.~\cite{beretta2018stochastic} argues that, among all models with the same number of parameters, models with a single ICC 
have the lowest MDL complexity. As MCMs are formed by the (independent) union of ICCs,
their MDL complexity is the sum of the complexity of each ICC, and is thus expected to be small.
We conjecture that MCMs are the models with 
the lowest MDL complexity among all spin models with the same number of parameters and the same rank. We verified that this is true for all models with up to $4$~spin variables (see App.~\ref{App:Complexity}).
Although this statement has yet to be proven in the general case, our belief in its likely correctness motivated our choice of the name ``minimally complex models''.

\subsection{Bayesian Model Selection} 
All these properties are good reasons for restricting BMS to the set of MCMs.
First, the factorization in Eq.~\eqref{factorization} shows that the best MCM for a given dataset provides sharp predictions on independencies between the variables modeled by different ICCs.
Second, the fact that the computation of the evidence in Eq.~\eqref{eq:evidenceMCM} is straightforward 
greatly simplifies the task of BMS, because comparisons between models can be done directly on the basis of their evidence without any parameter fitting. 
Finally, invariance under GTs of the family of MCMs ensures that the selected model does not depend on the basis 
in which the data is represented. 

Yet the number of MCMs is still astronomically large, even for moderate values of $n$ (growing roughly as $2^{n^2}$, see Fig.~\ref{Fig2:NbModels}).
To find the MCM with maximal evidence, we propose to divide the search in two steps. 
First, we search for the best model among all {\it Independent Models} (IM), which are MCMs with $r_a=1$ for all $a\in\mathcal{A}$ (see Fig.~\ref{Fig3:MCM_ex_GT}).
The maximization of the evidence over this family can be done by finding the set $\basis^*$ of the $n$ most biased independent operators in the dataset (see App.~\ref{Materials:proofIM}).
Although the number of IMs is also huge ($\sim 2^{n^2}$), finding the best IM with this procedure only takes of the order of $n\,2^n$ operations.
We developed a program performing the search for the best IM, which is available at Ref.~\cite{clelia_de_mulatier_2024_indep}.
Once the best IM~$\basis^*$ is found,
we reduce our search to the subset of MCMs that 
admit~$\basis^*$ as a preferred basis. 
This entails finding the optimal partition of the basis~$\basis^*$ into ICCs, which is the partition that maximizes the evidence. 
Searching exhaustively among all possible partitions of~$\basis^*$ can be done efficiently using a Gray code~\cite{ehrlich1973loopless, knuth2011art}; our program performing this search is available at Ref.~\cite{clelia_de_mulatier_2021_5578955}. 
The outcome of the overall algorithm can be represented as a factor graph with three layers (see Fig.~\ref{Fig4:SCOTUS}b), one connecting the original variables~$\spin$ to the preferred basis variables~$\basis^*$,
and the other connecting the latter to the ICCs 
of the best MCM$^*$ found in the~$\basis^*$ basis. 
Note that the number of IMs is not much smaller than the number of MCMs, whereas the number of MCMs with the same basis (denoted as MCM$^*$ in Fig.~\ref{Fig2:NbModels}) is much smaller. Indeed, the number of such models is given by the $n$-th Bell number, 
which grows slower than $(n/\log(n))^n$~\cite{berend2010improved}.
This makes exhaustive search possible for moderate system sizes ($n\lesssim 15$). Larger systems require further heuristics which will be discussed in section~\ref{Sec:LargeSystems}.

\section{MCM selection for small systems}
We first tested our approach on synthetic data. In particular, in 
App.~\ref{app:ToyModel} we discuss a toy example of a (non-noisy) binary dataset, in which one can appreciate how MCMs allow one to recover 
the simplest description of the data in terms of probabilistic boolean functions.
In the following, we discuss real-world examples.

\subsection{Voting data from the US Supreme Court~\cite{lee2015statistical}} Let us first illustrate our method with the analysis of voting data from the US Supreme Court database~\cite{USSCdatabase}. 
The dataset we consider is composed of the votes of the $n=9$ justices on $N=895$ cases debated during the second Rehnquist Court. 
As in Ref.~\cite{lee2015statistical}, voting outcomes were re-labeled using the political inclination (if known) of each debated case, such that, for each case, each judge casts a vote $s_i=-1$ if the decision is conservative-oriented, or $s_i=+1$ if it is liberal-oriented.
Ref.~\cite{lee2015statistical} showed that a fully connected pairwise model is able to explain higher-order features of the data. Ref.~\cite{Gresele} further analyzed the data within a more general scheme, showing that pairwise interactions are indeed prevalent in this dataset. 
The size of the system ($n=9$) is sufficiently small to perform the exhaustive search described in the previous section 
(the codes in~\cite{clelia_de_mulatier_2024_indep, clelia_de_mulatier_2021_5578955} take together less than a second to find the best MCM on a laptop).

\begin{figure}
    \centering
    \includegraphics[width=0.95\linewidth]{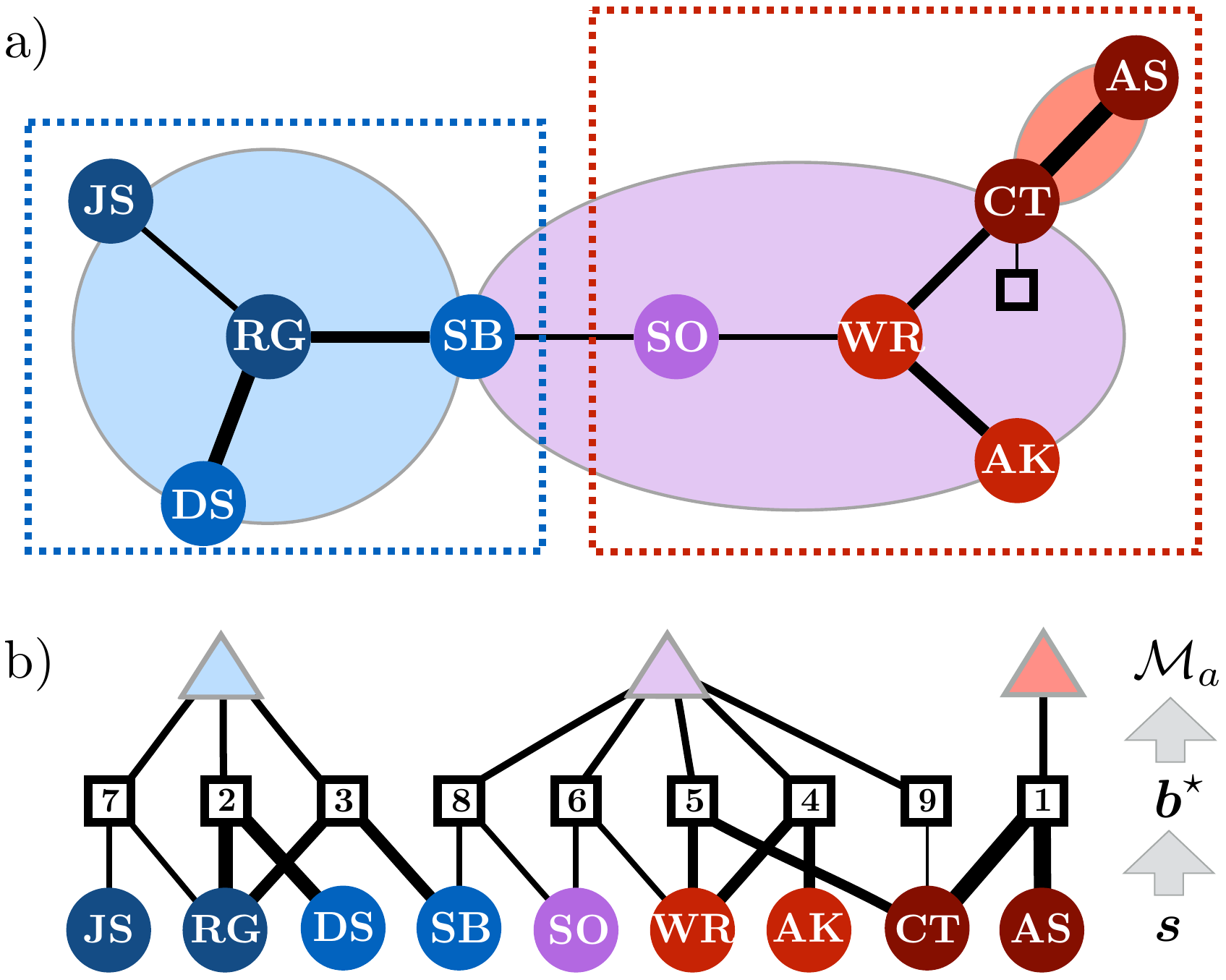}
\caption{
{\bf Analysis of the US Supreme Court Data.} 
Justices are represented by circles labeled by their initials:
Ruth Bader Ginsburg (RG), John P. Stevens (JS), David Souter (DS), Stephen Breyer (SB), Sandra Day O'Connor (SO), William Rehnquist (WR), Anthony Kennedy (AK), Clarence Thomas (CT), Antonin Scalia (AS). The colors represent their political orientation taken from Ref.~\cite{lee2015statistical}: dark red indicates the most conservative-oriented justices and dark blue the most liberal-oriented justices.
{\bf a) Best MCM, represented in the original basis variables} (i.e., the justices' votes).
The best IM is composed of $8$ pairwise interactions, represented by links between nodes with width proportional to their respective strength, and $1$ single-body interaction represented by a square on CT.
The strongest interaction has $\langle s_{\rm CT}s_{\rm AS}\rangle\simeq 0.86$,
whereas the weakest has $\langle s_{\rm CT}\rangle\simeq -0.45$.
The three large circles identify the partition of these interactions into the ICCs of the best MCM (among all).
The red and blue dotted squares indicate the partition of the judges that corresponds to the best MCM among those that have the original basis of the data as a preferred basis.
{\bf b) Factor graph representation of the best MCM.} Spin variables $\spin$ are represented by circles. The model selection procedure first identifies the best basis $\basis^*$, whose independent operators are denoted by squares, and then the best clustering of these operators into ICCs $\M_a$, denoted by triangles.
Each square in the second layer corresponds to one of the interactions of the best IM represented in Panel~a, numbered from the strongest interaction (1) to the weakest (9).
}
\label{Fig4:SCOTUS}
\end{figure}

The best IM, found using the exhaustive search in Ref.~\cite{clelia_de_mulatier_2024_indep}, is displayed in Fig.~\ref{Fig4:SCOTUS}.a. It has a very large log-evidence ($\log P(\hat s|\M)=-3327$) compared to the IM composed of $9$ single spin operators (for which $\log P(\hat s|\M)= -5258$).
Interestingly, we found that the $9$ most biased independent operators of the data are all single- or two-body, although the selection procedure takes into account interactions of all orders. 
This confirms the prevalence of low-order interactions in the system.
Moreover, these $9$ interactions alone account for more than $80\%$ of the information captured by the fully connected pairwise model with $45$ interactions, 
and therefore they must correspond to meaningful patterns of the data.
With the exception of the single spin interaction on CT, all these interactions are pairwise. They connect justices with similar political orientation~\cite{lee2015statistical} and define a network that spans the entire political spectrum.  
Interestingly, the strength of the interactions increases towards the extremes of the political spectrum and is weaker at the center.
Without the single spin interaction, which is much weaker than the other interactions, the systems would be invariant under spin reversal ($s_i\longrightarrow -s_i$). The single spin interaction on CT introduces a bias towards conservative votes.

The best MCM can be found from the best IM by using algorithms for generating all possible partitions of a set~\cite{ehrlich1973loopless, knuth2011art, generatingpartitions} (as implemented in Ref.~\cite{clelia_de_mulatier_2021_5578955}). There are $115\,975$ different MCMs that can be generated from a given IM with $n=9$. 
For the US Supreme Court data, the best MCM is composed of three ICCs (see Fig.~\ref{Fig4:SCOTUS}).
The first consists solely of the interaction between AS and CT, suggesting that AS's voting patterns 
conditioned on that of CT is mostly independent of all the others. The second ICC groups the three interactions between judges on the liberal side of the political spectrum, suggesting that, conditioned on the vote of SB, the voting patterns of RG, JS, and DS are mostly independent of the votes of the other judges. Finally, the third ICC groups interactions on the conservative-center side of the political spectrum.
We refrain from discussions on the political science implications of these results. Our aim is to underline how extracting the best MCM from the data can lead to 
interesting hypotheses on the voting behavior of the judges, grounded in the data.

For comparison, Fig.~\ref{Fig4:SCOTUS}.a also shows the best MCM among those that have the original basis of the data as a preferred basis (i.e., with $\basis=\spin$).
This MCM divides the court into two independent components that happen to match the justices' political orientations.
Yet the evidence of this model ($\log P(\hat s|\M)=-3300.4$) is considerably smaller than that of the MCM built from the best IM $\basis^*$ ($\log P(\hat s|\M)=-3154.4$).
The evidence of the best MCM in the original basis is actually smaller than the evidence of most MCMs of rank $9$ 
in the $\basis^{*}$~basis, which highlights the advantage of choosing the best IM as a basis for searching for the best MCM (see Appendix 
Fig.~\ref{fig:Distrib_log_evidence} for more details). 
For comparison with a fully connected pairwise model, the best MCM identified by our algorithm has fewer parameters ($K_{\rm MCM}=(2^5-1)+(2^3-1)+1=39$, while $K_{\rm pair}=9\times 10/2=45$).

The log-evidence of a model provides an estimate of the achievable level of compression of a dataset. 
For instance, the best MCM found in the original basis allows us to write the data with an average of $5.32$ bits/datapoint, while the best MCM overall can encode the data 
with an average of $5.08$ bits/datapoint, achieving respectively a compression of $40.9\%$ and $43.5\%$ of the original dataset.
The optimal compression of a dataset gives insights into its complexity, 
and the absolute value of the log-evidence of the model achieving such compression can be interpreted as the complexity of the data~\cite{rissanen1986stochastic, myung2000counting} (see also discussion leading to Eq.~\eqref{eq:MDL}).
In general, the complexity of a dataset is challenging to measure.
However, as illustrated here, MCMs can provide a feasible  
and efficient way of estimating the complexity of datasets, thanks to how easy 
it is to compute their evidence.

\section{MCM selection, heuristics for large systems}
\label{Sec:LargeSystems}
When $n$ is large and an exhaustive search is unfeasible, we propose the following heuristics. 
In order to find the best IM $\basis^{\star}$, we start from an initial basis $\basis$ (e.g., $\basis=\spin$), and we build all interactions up to a chosen order $k$. Among these interactions, we identify the set $\basis'$ of $n$ independent operators that are maximally biased, and replace $\basis$ by $\basis'$. We repeat this procedure until we find $\basis'=\basis$. 
This final basis is our best choice for the IM $\basis^{\star}$.
Although the exploration of the space of IM is limited by the choice of $k$, the iteration of this procedure is able, in principle, to explore the space of operators to any order. Our program implementing this algorithm is also available in Ref.~\cite{clelia_de_mulatier_2024_indep}.

Next, we apply a hierarchical merging procedure to find the optimal MCM. We start from the IM based on the basis operators $\basis^*$ identified above, which is an MCM with $n$ ICCs of rank $r_a=1$. 
We then compare all possible merging of two ICCs to identify the pair that yields the maximal increase of the evidence (if any). We merge the corresponding ICCs, and repeat this procedure until the evidence can't be increased by merging anymore.
This procedure finds the MCM that achieves the maximal value of the evidence along the hierarchical merging trajectory, 
as the number of ICCs varies from $n$ to $1$. Our program implementing this algorithm is available in Ref.~\cite{Greedy_Algo}. In the following sections, we apply this algorithm to several datasets, performing the iterative search for the best basis at the $k=4^{\rm th}$ order.
Detailed discussion about the performance of our heuristic algorithm is beyond the scope of the present work, which is focused on introducing MCMs.
However, some general observations to this effect are presented in 
App.~\ref{app:HeuristicAlgo:comments}.

\subsection{Big Five Personality Test~\cite{goldberg1992development}}
\begin{figure}
    \centering
    \includegraphics[width=\linewidth]{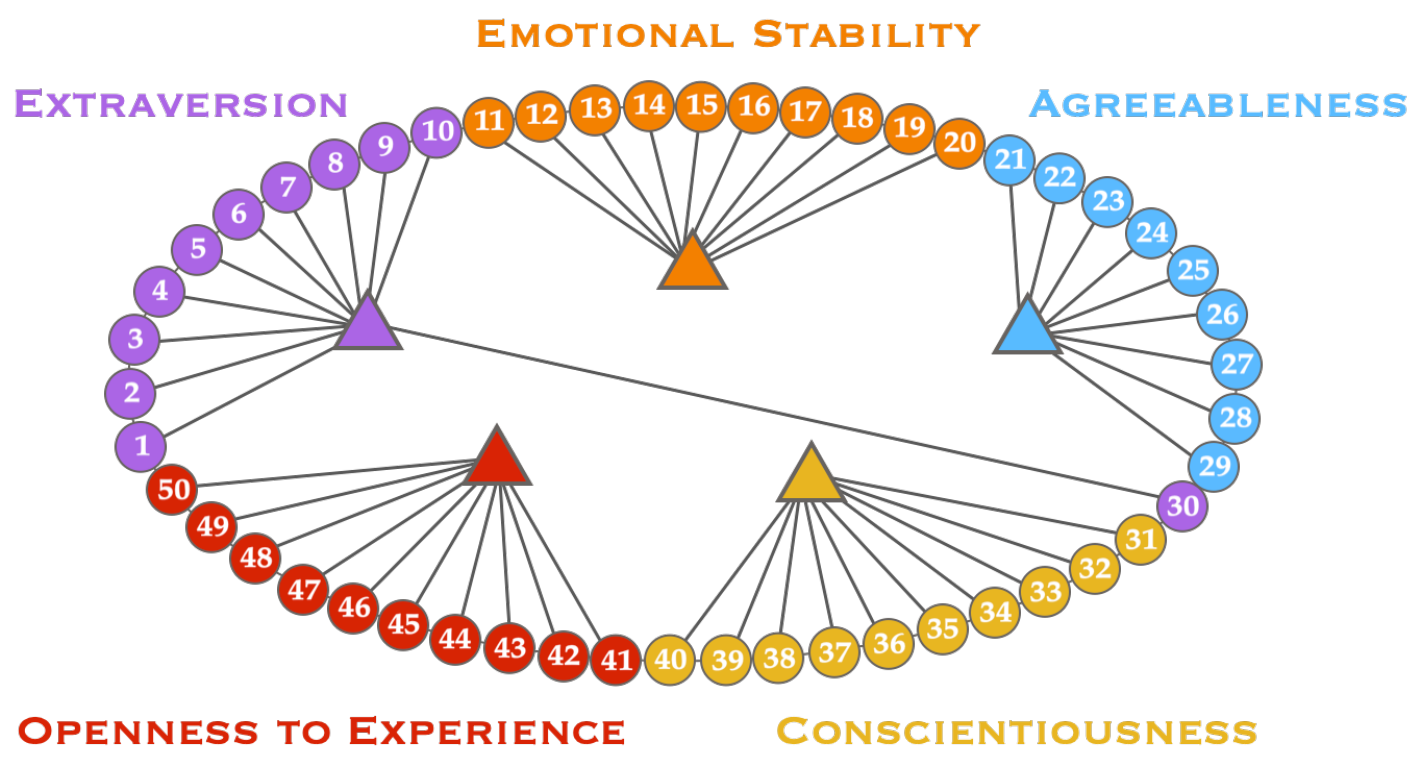}
\caption{ 
{\bf Best MCM found for the  Big Five Personality Test dataset.}
The numbered circles are the 50 questions ordered  
as in the original dataset of Ref.~\cite{Big5data}, in which questions were grouped by traits.
This is not a factor graph, as the best MCM found in the best basis happens to be the same model as the one displayed here in the original basis (see Fig.~\ref{fig:big5:Best_MCM}
for the factor graph representation in the best basis).
According to this result, the statement~30, ``{\it I make people feel at ease}'', is better associated to Extraversion than to Agreeableness.
}
\label{Fig5:Big5}
\end{figure}

Figure~\ref{Fig5:Big5} reports the resulting MCM for the Big Five Personality Test \cite{goldberg1992development}. The test consists of $n=50$ questions designed to probe the personality of individuals along five dimensions, that have been 
suggested as the main traits describing personality. These traits are extraversion, emotional stability, agreeableness, conscientiousness, and openness to experience.
Each trait is evaluated from the answers to ten questions, on a scale of one (disagree) to five (agree), that can be either positively or negatively associated with the trait.
For example, agreeableness is probed by questions such as {\it ``I sympathise with others' feelings''} or {\it ``I insult people''}. 
A dataset with $N=1\,013\,558$ samples was obtained from Ref.~\cite{Big5data}, to which we refer for more details. 
We converted each answer in binary format, depending on whether it was  
positively or negatively associated with the trait with respect to the average score across the whole sample. For this dataset, inference within the family of MCMs can reveal whether or not the data confirms the hypothesis that these questions probe the respondent's personality along five dimensions, 
and how well 
these dimensions are associated with the questions of the test.

\begin{figure*}
    \centering
    \includegraphics[width=0.78\linewidth]{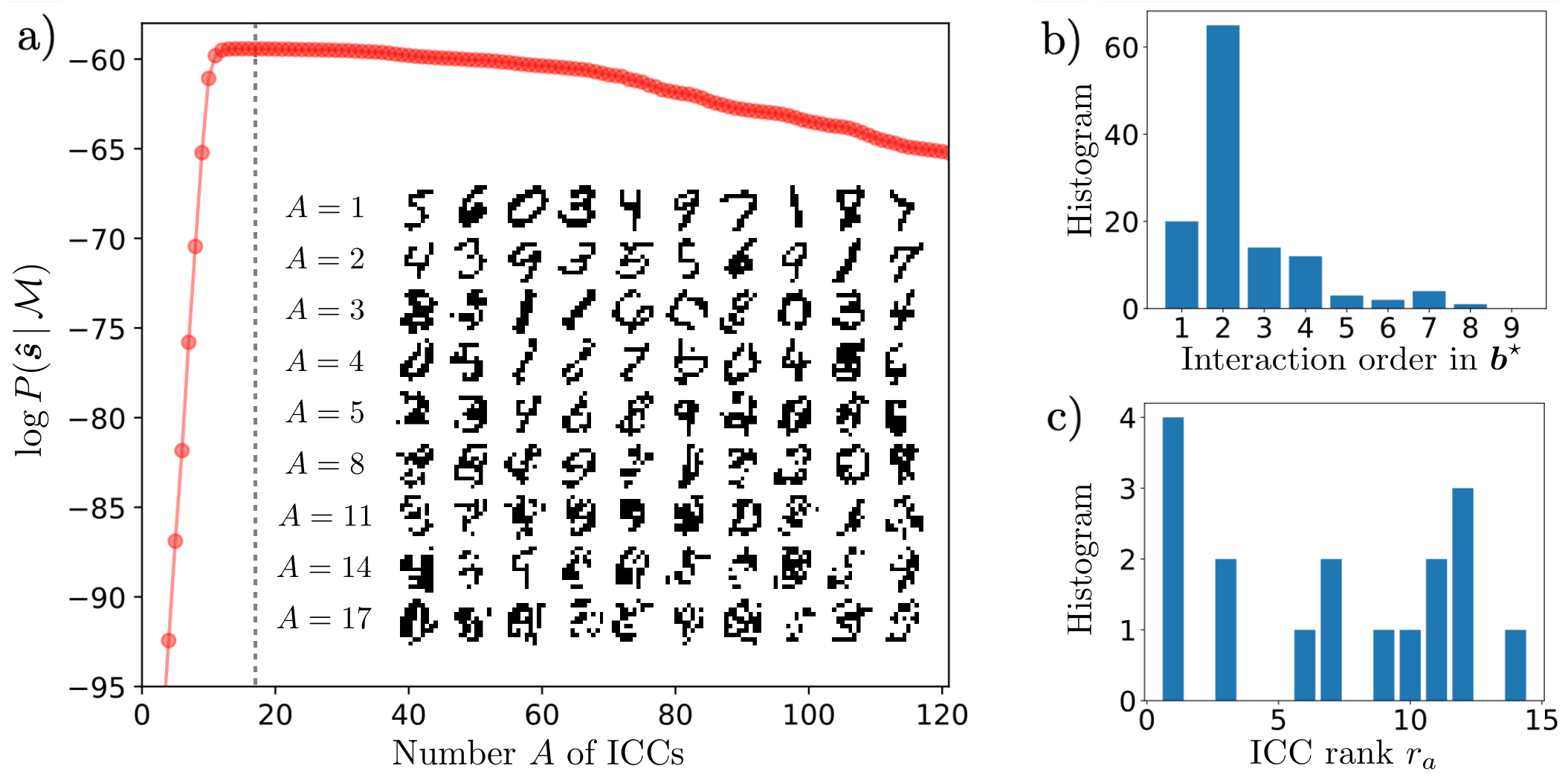}
\caption{ 
{\bf a)}
Log-evidence (in bits/datapoint) of the selected MCM as a function of the number $A=|\mathcal{A}|$ of ICCs 
during the merging procedure. The maximum is attained at $A=17$ (see vertical dashed line). 
{\bf b)}~Histogram of the order of the interactions in the best IM $\basis^{\star}$ found by the iterative search at the $4^{\rm th}$ order. There is a significant number of interactions that are of order higher than pairwise.
{\bf c)} Histogram of the rank $r_a$ (``size'') of the $17$ ICCs of the best MCM found with the merging procedure.
}
\label{Fig6:MNIST}
\end{figure*}

We applied the heuristic approach previously described~\footnote{Due to the large number of samples ($N\sim 10^7$), the best basis search~\cite{clelia_de_mulatier_2024_indep} to order $k=4$ took about 3~hours on a laptop and the greedy search~\cite{Greedy_Algo} for the best MCM$^*$ took a minute.}. 
Figure~\ref{Fig5:Big5} shows the best MCM we found overall, displayed in the original basis variables because of its simple visualization in this case. 
The factor graph representation of this model in the best basis can be found in Appendix Fig.~\ref{fig:big5:Best_MCM}. 
Here as well, the best basis $\basis^{\star}$ contains exclusively single-body and two-body operators. 
Except for one~\footnote{We found a significant pairwise interaction between question 3 {\it ``I feel comfortable around people''} that probes Extraversion and question 30 {\it ``I make people feel at ease''}, which probes Agreeableness.}, 
all these interactions are confined to questions relative to the same trait.
The best MCM also recovers the partition of the 50~questions along the 5~traits, 
except for question 30, ``{\it I make people feel at ease}'', 
that correlates better with Extraversion than Agreeableness.
The difference in log-evidence per datapoint is small, with 
$-35.835$~bits per datapoint
for the standard Big-Five MCM and $-35.802$ bits per datapoint
for the best MCM found. 
However, in view of the large number of datapoints, the different of log-evidence for the whole dataset is of the order of $10^{5}$.

A further analysis of the data showed that the Agreeableness trait is the least distinct relatively to the other traits.
For instance, moving question 23 ``{\it I insult people}'' from the Agreeableness ICC to the Emotional Stability ICC 
or to the Conscientiousness ICC 
doesn't significantly decrease the log-evidence (down respectively to $-35.845$ and $-35.857$ bits per datapoint -- 
see Fig.~\ref{fig:big5:Distrib_log_evidence} in Appendix),
suggesting that this question is also informative of these two other traits. 
In contrast, the questions probing Extraversion are mostly informative of that specific trait (see Fig.~\ref{fig:big5:Distrib_log_evidence}).
Moreover, the best basis analysis also identified the strongest pairwise dependencies to be within the Agreeableness ICC (see Fig.~\ref{fig:big5:Best_MCM}), 
namely, between questions 22 ({\it I’m interested in people}) and 27 ({\it I am not really interested in others}), 27 and
25 ({\it I am not interested in other people’s problems}), and 24 and 26. 
These strong dependencies contribute to the small value of log-evidence per variable in the Agreeableness ICC of $0.58$ bits per variable per datapoint;
in comparison, the other ICCs have values varying between $0.72$ and $0.76$.
This type of information could be used to improve the design of the test.

Overall, our analysis confirms that the questions of the Big Five Personality Test accurately probe the desired five independent traits~\cite{goldberg1992development}.
The analysis also identifies strong dependencies between questions of the same traits, 
as well as questions that are similarly informative of multiple traits. 
This section illustrates how the log-evidence of MCMs can be used 
to assess the quality of surveys such as personality tests and identify possible weaknesses, or inform future improved versions of the test.

\subsection{Modeling the MNIST database~\cite{MNIST}}
Finally, we performed BMS with MCMs on the MNIST database~\cite{MNIST}, which is composed of $N=60\,000$ images of hand-written digits. In order to reduce the data to a manageable size for our computational resources~\footnote{All calculations were performed on a laptop computer. For computational efficiency, our codes also use bitwise operations on 128-bit integers. This however limits the size of the systems we can handle to $n\le 128$.}, we coarse-grained it in cells of $2\times 2$ pixels, that were converted into binary values by applying a threshold~\footnote{If the sum of the gray levels of the four pixels exceeds 400, then the coarse-grained pixel is assigned the value ``1'', otherwise it is assigned ``0''.}. We focus on a central zone of $n=11\times 11=121$ pixels. Fig.~\ref{Fig6:MNIST}.a shows the evidence as a function of the number $A=|\mathcal{A}|$ of ICCs at different stages of the hierarchical merging procedure. 
A maximum is achieved at $A=17$; the structure of this MCM is rather complex, with a basis formed of several high-order interactions, which are grouped in relatively large ICCs (see Fig.~\ref{Fig6:MNIST}.b and c). 
We found that the elements of the best basis didn't change significantly when running the algorithm up to order $k=3$ or $4$ 
(see Fig.~\ref{fig:MNIST:SI} in Appendix). 
This suggests that the heuristic basis-search algorithm was able to quickly explore most of the relevant high-order basis operators, even at relatively small values of~$k$.

The aim of this section is to test the efficiency of the inferred MCMs in generalization.
For this purpose, we sample digits from the maximum likelihood distribution Eq.~\eqref{eq:MCM_P_s:maxL} for the MCMs found at different values of $A$ along the hierarchical merging procedure. 
Sampling from an MCM is remarkably cheap to perform, 
as it only requires drawing $A$ independent samples from the original data
and computing a few binary vector-matrix multiplications  
(see App.~\ref{Materials:Sampling}
--- our sampling code is available in~\cite{Greedy_Algo}).
For $A=1$ this procedure amounts to sampling 
digits directly from the original dataset. For $A>1$, sampling generates new patterns, as shown in the inset of Fig.~\ref{Fig6:MNIST} a). Although the sampled images have some structure, 
their resemblance to digits fades away as $A$ increases.

\section{Codes and linear algebra representation of spin models}

All the codes used for the data analyses presented in this paper are available in Ref.~\cite{clelia_de_mulatier_2024_indep, clelia_de_mulatier_2021_5578955, Greedy_Algo}.
Notably, our algorithms use a binary linear algebra representation of spin models and gauge transformations which is described in App.~\ref{Materials:Linear_Algebra}.
This representation allows us to significantly speed up the codes and 
provides a convenient mathematical framework for working with spin models.

\section{Discussion} 
Finding the model that best describes a dataset within Bayesian model selection is a daunting task. 
Here, we show that this task is considerably simplified when the selection is restricted to the family of Minimally Complex Models (MCMs). 
BMS within MCMs probes interactions of arbitrary order and can reveal the presence of high-order interactions or confirm that low-order interactions are prevalent in a dataset.
Yet, these models are simple in terms of their information-theoretic complexity, as defined by the Minimum Description Length principle~\cite{rissanen1986stochastic}.
This means that they are expected to provide robust predictions on dependencies in the data.
In particular, MCMs disentangle the structure of statistical dependencies by modeling data with independent model components:
the best MCM for a given dataset provides the factorization of the state probability distribution 
that best matches the statistical structure of the data.
More precisely, in some representations, the best MCM corresponds to a partition of the variables into groups, where variables are more correlated within groups than across groups (similarly to a community structure).
This is illustrated in several examples.

Thanks to this factorization property, 
future works in the modeling of very large systems could use MCMs as a pre-processing step for dividing the modeling problem into smaller sub-problems that can be analyzed in more detail. 
Similarly, MCMs provide a notion of locality that could be used in the context of coarse-graining or data renormalization procedures~\cite{meshulam2019coarse} going beyond pairwise correlations.
Moreover, MCMs allow for modeling data with interactions of arbitrary order and thus open up new ways to infer high-order interaction structures from data. 
This contributes to advancing research in the physics of high-order interactions in complex systems~\cite{rosas2019quantifying, battiston2021physics, ruggeri2023community}.
Finally, the invariance of the family of MCMs under Gauge Transformations (GTs)
ensures that the inference process is independent of how data is represented~\footnote{Consider, for a purely illustrative purpose, an idealized problem of inference in a gene regulatory network. 
Assume that whether gene $i$ is expressed ($s_i=+1$) or not ($s_i=-1$) depends on whether its $t_i$ transcription regulators are bound ($\sigma_{i,a}=+1$) or not ($\sigma_{i,a}=-1$) to the regulator binding region, i.e. that $s_i=f_i(\sigma_{i,1},\ldots,\sigma_{i,t_i})$ is a boolean function of the $\sigma_{i,a}$'s. If the relation between $\spin$ and $\boldmath{\sigma}$ is a GT, then our approach ensures that 
inference of the gene regulatory network based on a dataset $\dataset$ of gene expression should give the same results as inference based on a dataset $\boldsymbol{\hat{\sigma}}$ of binding on regulatory regions.}. This is not true when model selection is restricted to a family of models that is not invariant under GTs, such as the family of pairwise models. In this case, it may be hard to say whether the structure of the inferred model reflects the statistical dependencies of the data or the constraints imposed by 
the considered family of models. 

Our approach departs from the literature on graphical model reconstruction~\cite{montanari2009graphical,ravikumar2010high,barton2016ace,vuffray2016interaction} in important ways.
Graphical model reconstruction aims at retrieving the model from which a given data could have been generated,
by identifying which interactions among a pre-assigned set (e.g. pairwise) are present.
In this work, we don't make any assumptions on the interactions present 
in the models, 
as our approach takes into account all possible (low- and high-order) correlation patterns in the data when searching for the best model.
Even more importantly, we are not interested in reconstructing and fitting 
a specific set of interactions from which the data could have been generated.
Instead, we aim at finding the most likely {\it simple} model that can inform us on true structure of dependencies in the data. 
Another critical aspect in graphical model reconstruction is parameter estimation, for which different approximate algorithms have been proposed~\cite{nguyen2017inverse}. 
We show that there is no need to infer model parameters in order to perform model selection among MCMs, which is a huge computational advantage. 
Moreover, MCMs can be directly compared on the basis of their evidence (marginal likelihood), 
which allows to balance goodness-of-fit and model complexity. 
In contrast, the complexity of graphical models is 
difficult to compute, 
which calls for adopting regularization methods~\cite{ravikumar2010high, nguyen2017inverse} that don't always have a clear information-theoretic foundation. 
Finally, the number of MCMs that can be compared with BMS is much larger than the number of possible pairwise models 
(see Fig.~\ref{Fig2:NbModels}), covering more widely the space of all statistical structures of data.

As mentioned previously, models of the same class can be seen as different representations of the same abstract model.
In our approach to BMS of MCMs, we aim to find the best model for the data independently of its representation; in other words, we aim to find the best abstract model.
This approach to model selection that is independent of the representation of the data 
may pave the way toward a better understanding of the concept of abstraction.
Put differently, our results show that the constraint on model simplicity {\em i)} defines 
an extremely vast landscape 
of possible data structures that {\em ii)} is endowed with a rich relational organization and {\em iii)} is easy to navigate by Bayesian model selection algorithms. These remarkable facts make MCMs a good candidate for the ``more sophisticated forms of knowledge representation'' that Tenenbaum {\em et al.}~\cite{tenenbaum2011grow} argue ``must underlie the probabilistic generative models needed for Bayesian cognition''.

In summary, this paper offers a novel perspective on statistical inference of high-dimensional data by focusing on simple representations of data. 
The computational efficiency of performing BMS among MCMs, their ability to uncover robust structures of dependencies in the data, and their interpretability in terms of groups of highly correlated variables make MCMs a very good tool for fast and interpretable data analysis. 
MCMs open up new ways of tackling challenging statistical modeling problems, such as modeling very large systems or identifying significant high-order patterns in data.
Further extensions beyond binary variables, as well as the development of 
efficient optimization algorithms 
to find the MCM with maximal evidence for large systems, are
promising avenues of research. 

\begin{acknowledgments}
We gratefully acknowledge Edward Lee, Chase Broedersz, and William Bialek for sharing the data of~\cite{lee2015statistical}. 
We are grateful to Iacopo Mastromatteo, Vijay Balasubramanian, Yasser Roudi, and Eugenio Piasini for insightful discussions.
\end{acknowledgments}

\appendix

\section{\bf Spin operators and spin models}
\label{app:spinModels}
A {\em spin model} is a probabilistic model that describes the state of a system of binary variables, called {\em spins}. 
The model assumes the existence of interactions between the spins that constrain the states accessible to the system. As no spatial organization is assumed,
interactions can be of arbitrary range and arbitrary order (i.e., between two, three, or more spins). 
Spin models can be seen as a generalization of the Ising model to include long-range and high-order interactions.\\

\paragraph*{\bf Spin operators.}
To mathematically define a spin model, each interaction in the model is associated with a {\em spin operator}. 
Consider a system of $n$ spin variables, $\spin = (s_1 ,\,\cdots, s_n )$, that take random binary values $s_i = \pm 1$. A {\em spin operator} $\phi^{\vecmu}(\spin)$ associated with an interaction labeled by $\vecmu$ is defined as the product of the spin $s_i$ involved in the interaction:
\begin{equation}\label{eq:phi}
    \phi^{\vecmu}(\spin) = \prod_{i=1}^{n}\, s_i^{\,\mu_i}\,.
\end{equation}
Here, $\vecmu$ is an $n$-dimensional binary vector encoding which spin is involved in the interaction, i.e., its $i$-th element is $\mu_i=1$ if spin $s_i$ is included in the interaction and $\mu_i=0$ otherwise. Thus the spin $s_i$ remaining in the product of Eq.~\eqref{eq:phi} are the ones for which $\vecmu_i=1$.
For example, in the model in the bottom left corner of Fig.~\ref{fig:SI:MCM_GT}, 
the 3-body interactions between the variables $s_4$, $s_5$ and $s_6$ is encoded by the binary vector $\vecmu = (0,0,0,1,1,1)$ and associated with the spin operator $\phi^{\vecmu}(\spin)=s_4s_5s_6$; the pairwise interaction between $s_2$ and $s_3$ is encoded by $\vecmu = (0,1,1,0,0,0)$ and associated with $\phi^{\vecmu}(\spin)=s_2s_3$; and the field interaction on $s_2$ (represented by a black dot) is encoded by $\vecmu = (0,1,0,0,0,0)$ and associated with $\phi^{\vecmu}(\spin)=s_2$. 
The number of possible spin operators is equal to the number of non-null binary vector $\vecmu$, which is $2^n-1$ in an $n$-spin system.\\

\paragraph*{\bf Properties of spin operators.}
With the addition of the identity operator, $\phi^{\boldsymbol{0}}(\spin)=1$ for all $\spin$, the set of all spin operators, denoted $\Omega_n = \{\phi^{\vecmu}(\spin)\,|\,\vecmu\in\{0,1\}^n\}$, with the product operation forms a commutative 
group. Indeed, the square of any spin variable $s_i$, and therefore of any operator, is equal to the identity $\phi^{\boldsymbol{0}}(\spin)=1$. Besides, the product $\phi^{\vecmu}(\spin)\phi^{\vecnu}(\spin)=\phi^{\vecmu\oplus\boldsymbol{\nu}}(\spin)$ of any two spin operators is also a spin operator, where $\oplus$ denotes the component-wise 
XOR Boolean operator (i.e., sum modulo 2) applied to the two binary vectors $\vecmu$ and $\vecnu$.

The set of operators also forms a complete and orthogonal basis for all functions of $n$ binary variables, as
\begin{align}\label{eq:completeness:1}
    \sum_{\spin\in\{\pm 1\}^n}\phi^{\vecmu}(\spin)\,\phi^{\vecnu}(\spin)=2^n\,\delta_{\vecmu,\vecnu}\\
    {\rm and}\;\;\sum_{\vecmu=0}^{2^n-1}\phi^{\vecmu}(\spin)\,\phi^{\vecmu}(\spin')=2^n\,\delta_{\spin,\spin'}. \label{eq:completeness:2}
\end{align}
Hence any function $F(\spin)=\sum_{\vecmu\ge 0} f_{\vecmu}\phi^{\vecmu}(\spin)$ defined on the spin configurations can be represented as a linear combination of spin operators, with coefficients given by $f_{\vecmu}=2^{-n}\sum_{\spin}\phi^{\vecmu}(\spin)F(\spin)$. These completeness relations hold also on the sub-space defined by a subset of the spins.\\ 

\paragraph*{\bf Spin models.} A {\em spin model} is a maximum entropy model, whose Hamiltonian 
is a linear combination of the elements of a set $\M$ of spin operators, $\M\subseteq\Omega_n\backslash\{\phi^{\boldsymbol{0}}\}$.
The probability distribution over the spin variables $\spin$ under a spin model $\M$ is therefore: 
\begin{align}\label{app:def:spinmodel}
    P(\spin\,|\,\vecg,\,\M) = \frac{1}{Z_{\M}(\vecg)}\,\exp\bigg(\sum_{\vecmu\in\M}g_{\vecmu}\,\phi^{\vecmu}(\spin)\bigg)\nonumber\\
    \qquad{\rm with}\;\;
    Z_{\M}(\vecg)=\sum_{\spin}\,\exp\left({\sum_{\vecmu\in\M}g_{\vecmu}\,\phi^{\vecmu}(\spin)}\right)\,,
\end{align}
where $\vecg = \{g_{\vecmu},\, \vecmu\in\M\}$ is a vector of real parameters and where the partition function $Z_{\M}(\vecg)$ ensures normalization.
Each parameter $g_{\vecmu}$ modulates the strength of the interaction associated with the operator $\phi^{\vecmu}(\spin)$. Each spin model thus defines a parametric family of probability distributions.
In an $n$-spin system, there are $2^n-1$ different spin operators, hence there are $2^{2^n-1}$ different spin models that can be constructed. Each model is identified by the set $\M$ of spin operators that it contains.
For example, the second model of the first column in Fig.~\ref{fig:SI:MCM_GT} has $7$ interactions (three 1-body, three 2-body, and one 3-body) which are identified by the set of operators $\M=\{s_4, s_5, s_6, s_4s_5, s_4s_6, s_5s_6, s_4s_5s_6\}$ and which defines the spin models whose Hamiltonian is given by:
\begin{align}
    g_{\cdots100}\, s_4 + g_{\cdots010}\,s_5 + g_{\cdots010}\,s_6 + 
    g_{\cdots110}\,s_4s_5 + \nonumber\\
    g_{\cdots101}\,s_4s_6 + g_{\cdots011}\,s_5s_6 +
    g_{\cdots111}\,s_4s_5s_6\;.
\end{align}
Here we only wrote the term in the exponential of Eq.~\eqref{app:def:spinmodel}
and for clarity, we replace the three first spins ($s_1$ to $s_3$) with dots in the labels of the parameters.

\section{Gauge transformations}
\label{App:GT}
\paragraph*{\bf Set of independent operators.}
A set of operators is said {\it independent} if none of the operators of the set can be obtained as a product of other operators of the set. 
The sets $\{s_1,\,s_2,\, s_3\}$ and $\{s_1,\, s_1 s_2,\, s_1 s_2 s_3\}$ are two examples, 
whereas $\{s_1,\, s_2,\, s_1 s_2\}$ is not an independent set 
(as each operator can be obtained as a product of the two others). We recall that 
for a spin variables $s_i=\pm 1$
and therefore $(s_i)^{2} = 1$ always.

In an $n$-spin system, the set of the $n$ spin variables 
$\{s_1,\ldots,s_n\}$ is a set of independent operators. 
Besides all the operators of the system can be generated as products of these $n$ basis spin operators.  
As a result, in an $n$-spin system, a set of independent operators can have at maximum $n$ elements. The completeness relations~\eqref{eq:completeness:1} and~\eqref{eq:completeness:2} also hold on the sub-space defined by a set of independent operators.\\

\begin{figure}[h!]
\centering
    \includegraphics[width=.975\linewidth]{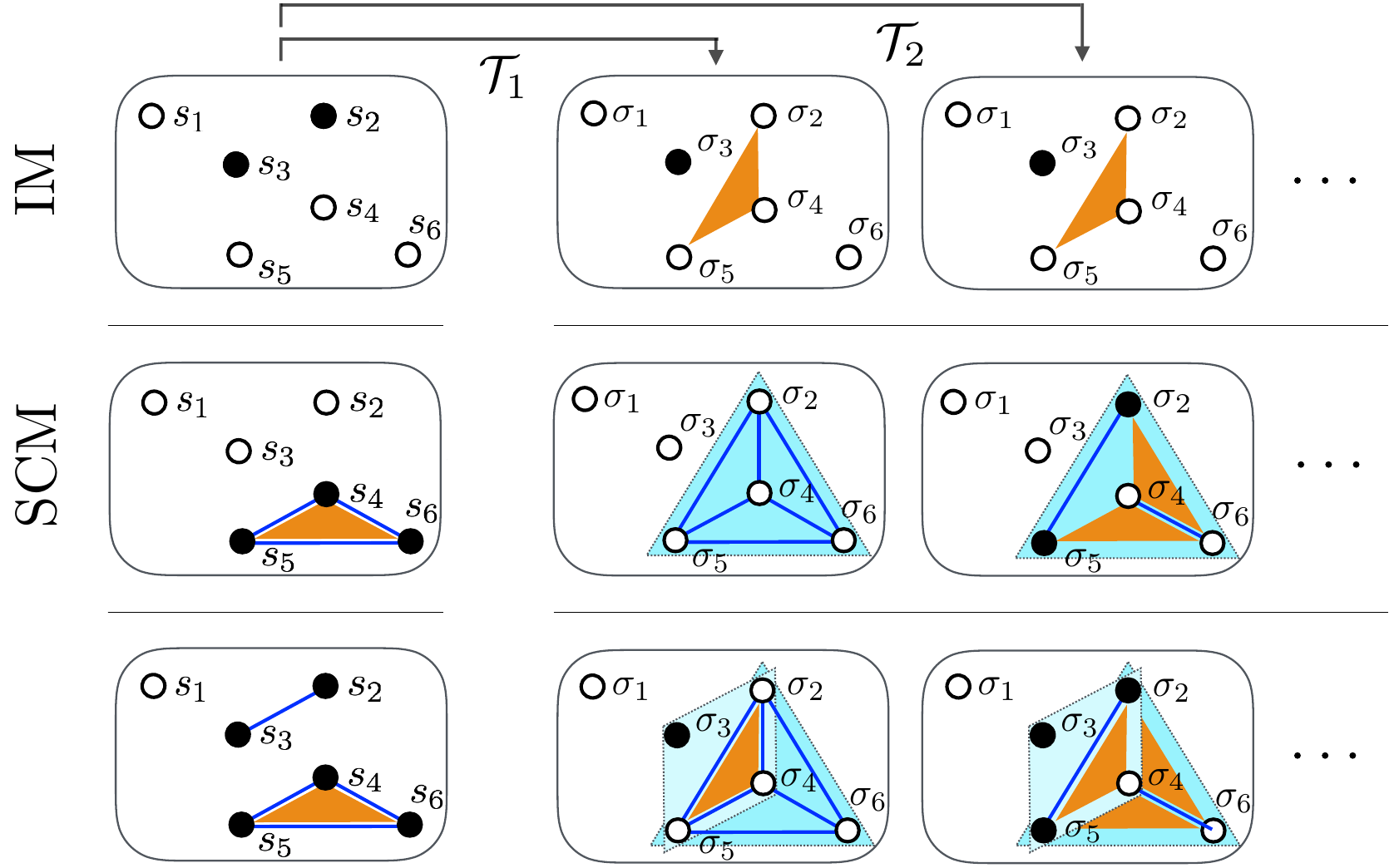}
    \vspace{3mm}\\
    \includegraphics[width=0.75\linewidth]{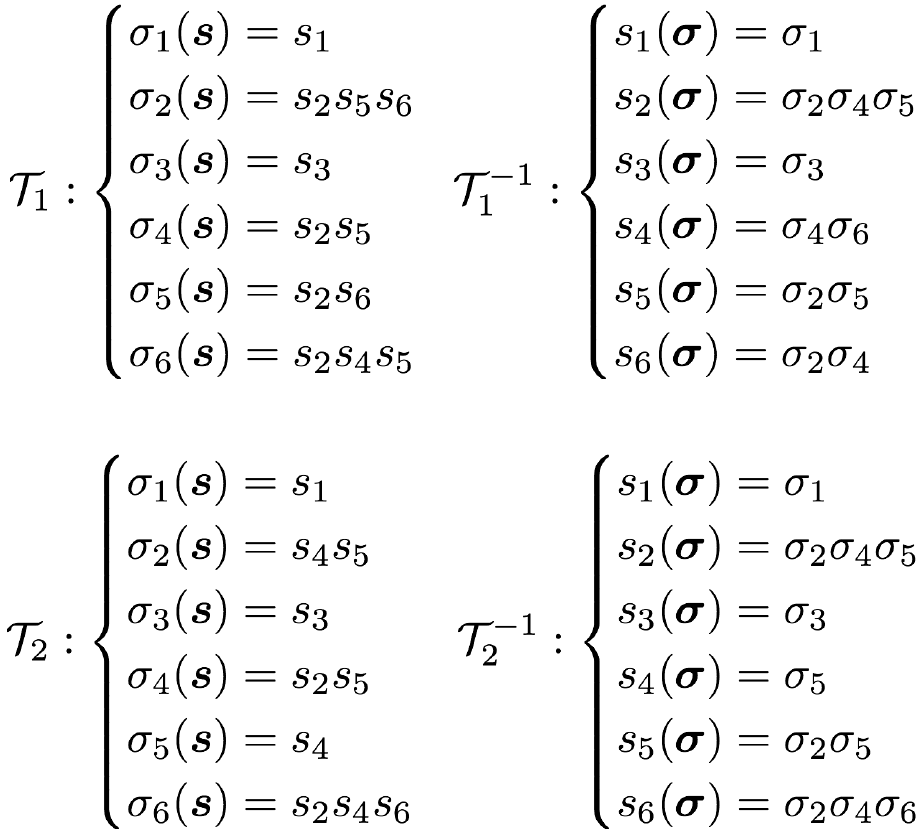}
\caption{(colors online)
    {\bf Additional information for Fig.~\ref{Fig3:MCM_ex_GT}:}
    Basis transformations $\mathcal{T}_1$ and $\mathcal{T}_2$ used to transform the models of the first column into the models of the 2nd and 3rd columns respectively. The inverse transformations (respectively, $\mathcal{T}_1^{-1}$ and $\mathcal{T}_2^{-1}$) are also provided.
    }
\label{fig:SI:MCM_GT}
\end{figure}

\paragraph*{\bf Gauge transformations.}
In an $n$-spin system $\spin = (s_1,\,\cdots,\,s_n)$, consider the set of $n$ independent operators $\basis(\spin) = (\phi^{\vecmu_1}(\spin),\, \cdots,\,\phi^{\vecmu_n}(\spin))$. The change of variables $\spin^\prime = \basis(\spin)$, in which each new variable $s_i^{\prime}$ takes the binary value $s_i^{\prime}=\phi^{\vecmu_i}(\spin)$,
defines a bijective map from the space of states (of the system) to itself.
Ref.~\cite{beretta2018stochastic} calls these transformations {\em gauge transformations} (GTs) and identifies them as being the automorphisms of the group $({\mathbb{Z}_2}^n, \cdot)$ (see Supplementary Materials of~\cite{beretta2018stochastic}). Permutations of the spins are special examples of GT. The transformations $\mathcal{T}_1$ and  $\mathcal{T}_2$ in Fig.~\ref{fig:SI:MCM_GT} are two other examples. 
As the transformation $\spin^\prime = \basis(\spin)$ is a bijection of the space of states, one can also define the inverse transformation: $\spin = \basis^{-1}(\spin^\prime)$. For example, Fig.~\ref{fig:SI:MCM_GT} gives the inverse of the transformations $\mathcal{T}_1$ and  $\mathcal{T}_2$.
In the following, we will denote $\mathcal{T}_{\basis}$ the GT defined by the set $\basis(\spin)$ of $n$ independent operators (i.e. $\mathcal{T}_{\basis}:\;\spin\rightarrow \spin^\prime = \basis(\spin)$) and $\mathcal{T}_{\basis}^{-1}$ its inverse transformation.
Appendix.~\ref{Materials:Linear_Algebra} gives information on how to use linear algebra to mathematically define GTs and obtain their inverse.

More generally, GTs can be thought of as a change of basis from the original spin representation to a new one. Under a GT, an operator is transformed into another operator and a spin model into another spin model.\\ 

\paragraph*{\bf Transformed operator.} 
Any spin operator $\phi^{\vecmu}(\spin)$ can be transformed under a gauge transformation $\mathcal{T}_{\basis}$ 
by simply re-expressing $\phi^{\vecmu}(\spin)$ in terms of the new basis variables $\spin^\prime=\basis(\spin)$. This can be done using the inverse gauge transformation $\spin = \basis^{-1}(\spin^\prime)$, which yields the transformed operators:
\begin{align}\label{App:eq:Phi_prime}
    \mathcal{T}_{\basis}[\,\phi^{\vecmu}\,](\spin') 
    \doteq \phi^{\vecmu}(\basis^{-1}(\spin'))\,.
\end{align}
One can easily check that the resulting expression for $\mathcal{T}_{\basis}[\,\phi^{\vecmu}\,](\spin')$ is still a spin operator but now expressed as a product of the transformed variables $s_i^{\prime}$. This means that there exists a binary vector $\vecmu^\prime$ such that:
\begin{align}\label{App:eq:Phi_prime:2}
    \phi^{\vecmu^{\prime}}(\spin') = \mathcal{T}_{\basis}[\,\phi^{\vecmu}\,](\spin')
    = \phi^{\vecmu}(\basis^{-1}(\spin'))\,.
\end{align}
Appendix~\ref{Materials:Linear_Algebra} gives the mathematical relation between $\vecmu^\prime$, $\vecmu$, and $\basis(\spin)$ (see Eq.~\eqref{Methods:GT:Operator}).\\

\paragraph*{\bf Transformed model.} The gauge transformation $\M^{\prime}$ of a model $\M$ is obtained by transforming each operator of $\M$:
\begin{align}\label{App:eq:M_prime}
    \M'=\mathcal{T}_{\basis}[\M] \doteq \{\mathcal{T}_{\basis}[\phi^{\vecmu}]\;|\; \phi^{\vecmu}\in\M\}\,.    
\end{align}
Because $\mathcal{T}_{\basis}$ is a bijection of the space of states, the probability of finding the system in the state $\spin$ in model $\M$ is the same as the probability of finding the system in the transformed state $\spin^\prime = \basis(\spin)$ in the transformed model $\M^\prime = \mathcal{T}_{\basis}[\M]$:
\begin{align}
\label{GT:def_on_model}
p(\spin\,|\,\vecg,\mathcal{M})
    &=p(\spin^{\prime}\,|\,\vecg^{\prime}, \mathcal{M}^{\prime})\,,\\
    &=p\left(\basis(\spin)\,|\,\mathcal{T}_{\basis}[\vecg], \mathcal{T}_{\basis}[\M]\right)\,,
\end{align}
where the vector of parameters $\vecg^\prime=\mathcal{T}_{\basis}[\vecg]$ is a permutation of the parameters $\vecg$ of $\M$ that is such that
the transformed operator $\phi^{\vecmu'}= \mathcal{T}_{\basis}[\phi^{\vecmu}]$ in $\M'$ is parameterized by the same parameter as the operator $\phi^{\vecmu}$ in $\M$.
Fig.~\ref{fig:SI:MCM_GT} shows examples of GTs of spin models.\\

\section{Complexity of MCMs}
\label{App:Complexity}

\begin{figure}[!ht]
\centering
\includegraphics[width=\linewidth]{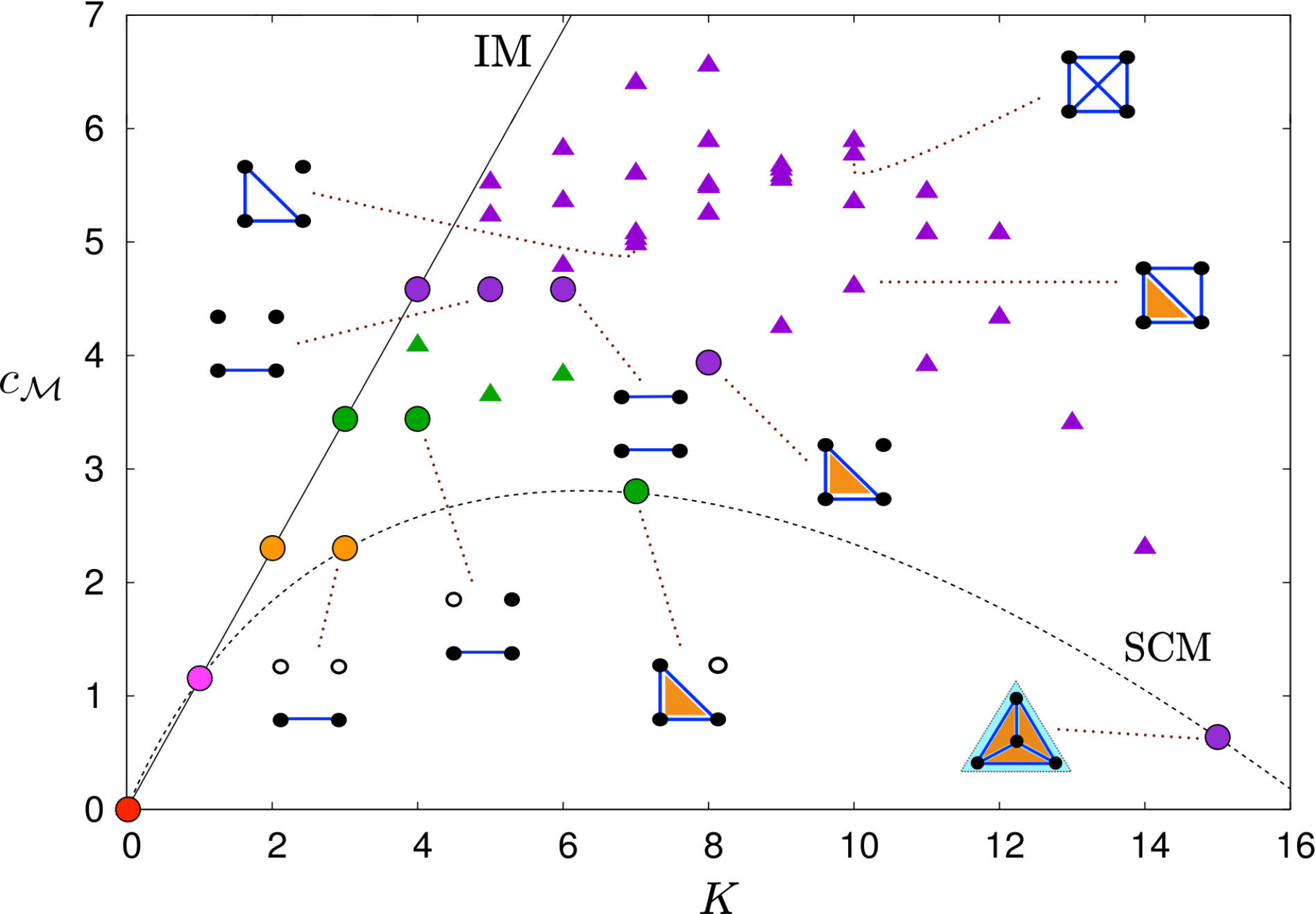}
\caption{(colors online)
    {\bf Complexity $c_{\mathcal{M}}$ of all spin models with $n=4$ spins} as a function of the number of parameters $K$ in the model -- adapted from 
    Ref.~\cite{beretta2018stochastic}. 
    Models of the same class have the same complexity. In the original figure, each class is represented by a triangle; here, we replaced the triangles by a circle for the classes of MCMs.
    Model representants of some of the classes are displayed using the same diagrammatic notation as in Fig.~\ref{Fig1:MCM_ex}.
    The complexity value of any model lies between two limit lines:
    one corresponds to the complexity of IM (solid line), the other to the complexity of SCM (dashed line).
    The colors indicate the number $r$ of independent variables in the model (i.e., the {\em rank} of the model): red for~0, pink for~1, green for~3 and violet for~4.
    Among all models with same number $K$~of parameters and same rank $r$, MCMs are the simplest.
    }
\label{fig:MCM_complexity}
\end{figure}

Assuming Jeffreys' prior over the model parameters, the {\em geometric complexity} of spin models in Eq.~\eqref{eq:MDL} is: 
\begin{align}\label{complexity}
    c_{\M} = \log \int d \vecg \sqrt{\det \FIM_{\M}(\vecg)}, 
\end{align}
where $\FIM_{\M}(\vecg)$ is the Fisher information matrix (FIM):
\begin{equation}\label{FIM}
    [\FIM_{\M}(\vecg)]_{\mu,\nu} = \partial_{g_\mu}\partial_{g_\nu}\left[\log Z_{\M}(\vecg)\right]\,.
\end{equation}
Ref.~\cite{beretta2018stochastic} shows that gauge transformations (GTs) leave the partition function $Z_{\M}(\vecg)$ of a spin model invariant up to permutations of its parameters. 
As a consequence, the FIM in Eq.~\eqref{FIM} and the geometric complexity $c_{\M}$ of spin models are both invariant under GTs.
This allows the classification of all spin models into equivalence classes characterized by the same complexity~$c_{\M}$.
Fig.~\ref{fig:MCM_complexity}, adapted from Ref.~\cite{beretta2018stochastic}, shows the complexity of all the classes of spin models on $n=4$ variables. 

The geometric complexity of spin models is generally difficult to compute due to the high-dimensional integral in Eq.~\eqref{complexity}. However, one can get a closed-form expression for MCMs.
As a consequence of the factorization property of MCMs in Eq.~\eqref{factorization}, the FIM of an MCM $\M = \cup_{a\in\mathcal{A}} \M_a$ is a block diagonal matrix, in which each block $\FIM_{\M_{a}}(\vecg_a)$
is the FIM of the ICC $\M_a$ reduced to the modeled $r_a$ spin variables. As a result, the determinant of the FIM for an MCM factorizes over the FIM of its ICCs, 
and the complexity of an MCM is then the sum of the complexity of its ICCs:
\begin{align}\label{SM:eq:MCM:Complexity}
    &c_{\M} =\sum_{a\in\mathcal{A}}  \,c_{\M_{a}}\,,\\
    &{\rm where}\;\; c_{\M_{a}} = \log \int d \vecg_a \sqrt{\det \FIM_{\M_a}(\vecg_a)}. \nonumber
\end{align}
Finally, the complexity of ICCs was computed in Ref.~\cite{beretta2018stochastic} (as the complexity of Sub-Complete Models):
\begin{align}\label{SM:eq:SCM:Complexity}
    c_{\M_a }
        \,=\,
        2^{r_a-1}\log\pi - \log\Gamma(2^{r_a-1})\,,
\end{align}
where $\Gamma$ is the gamma function. 
In Fig.~\ref{fig:MCM_complexity}, we indicated the classes of MCMs with a circle symbol, and one can see that, for $n=4$ or smaller, MCMs  
have the lowest complexity among all models with the same rank and the same number of parameters. This remains to be proven for larger values of $n$. 

For the special cases of Independent Models (IM) and Sub-Complete models (which are MCMs with a single ICC), Eq.~\eqref{SM:eq:MCM:Complexity} gives back the complexity values obtained by Ref.~\cite{beretta2018stochastic}, i.e., respectively, $c_{\M} = K\log \pi$ indicated by a solid line in Fig.~\ref{fig:MCM_complexity}, and the complexity in Eq.~\eqref{SM:eq:SCM:Complexity} indicated by a dashed line.

\section{Factorization of the probability distribution of an MCM}
\label{App:Factorization}
Consider an MCM $\M=\cup_{a\in\mathcal{A}} \M_a$ and one of its preferred bases $\basis(\spin)$. 
By definition of a {\it preferred basis}, $\basis(\spin)$
can be written as $\basis(\spin)=\cup_{a\in\mathcal{A}} \basis_a(\spin)$, where each $\basis_a(\spin)$ is a basis of the ICC $\M_a$. 
Expressing $\M$ in the basis $\basis(\spin)$ gives the transformed MCM $\M^\prime = \mathcal{T}_{\basis}[\M] = \cup_{a\in\mathcal{A}} \M_a^\prime$, in which the ICCs $\M_a^\prime = \mathcal{T}_{\basis_a}[\M_a]$
are now based on distinct basis variables~$\basis_a$.
Thus, the probability distribution of $\M^\prime$ (reduced to the sub-space defined by $\basis(\spin)$) is factorizable over its ICCs:
\begin{align}\label{method:eq:GT_MCM}
    p_{r}(\basis(\spin)\,|\,\vecg^\prime,\, \M^\prime)
    =
    \prod_{a\in\mathcal{A}} p_a(\basis_a(\spin)\,|\,\vecg_a^\prime,\, \M_a^\prime)\,,
\end{align}
where $\M_a^\prime=\mathcal{T}_{\basis_a}[\M_a]$ 
and where the parameters $\vecg_a^\prime$ of $\M_a^\prime$ are the permutations of the parameters $\vecg_a$ of $\M_a$ given by \eqref{method:eq:g_prime}. Here, each $p_a$ is a probability distribution over $r_a=|\basis_a|$ binary variables, and the model distribution $p_{r}$ is over $r=\sum_{a} r_a = |\basis|$ variables.

Note that the rank $r$ of an MCM can be smaller than $n$, in which case $\basis(\spin)$ is not a basis of the whole $n$-dimensional spin system, 
but only of the $r$-dimensional sub-system modeled by $\M$.
The basis $\basis(\spin)$ thus defines a gauge transformation for this sub-system (as used above), but not for the whole $n$-spin system. 
Completing $\basis(\spin)$ with $(n-r)$ independent operators to form a basis $\boldsymbol{B}(\spin)$ of the whole $n$-spin system, we can define a gauge transformation of the $n$-spin system that will transform $\M$ into the same model $\M^\prime$, i.e. such that $\mathcal{T}_{\boldsymbol{B}}[\M]=\mathcal{T}_{\basis}[\M]=\M^\prime$.
Using \eqref{GT:def_on_model} for the probability distribution of gauge transformed models, we can then write that:
\begin{align}
    p(\spin\,|\,\vecg,\, \M)
    =\frac{1}{2^{(n-r)}}\,
    p_{r}(\basis(\spin)\,|\,\vecg^\prime,\, \M^\prime)\,.
\end{align}
Combining this with \eqref{method:eq:GT_MCM} finally leads to the factorization in Eq.~\eqref{factorization}. 

\section{Maximum Likelihood and Evidence of MCMs}\label{Materials:Evidence}
\paragraph*{\bf Change of parametrization.}
Let us consider an ICC $\M_a$ with rank $r_a$ and a basis $\basis_a(\spin)=(\phi_1(\spin),\ldots,\phi_{r_a}(\spin))$ of $\M_a$. 
The probability distribution of the states of the system under $\M_a$ can be re-written in the basis of operators $\basis_a$ as:
\begin{equation}
\label{eq:ICC}
p(\spin\,|\,\vecg,\M_a)
    =\frac{1}{2^{(n-r_a)}}\, q\left(\basis_{a}(\spin)\right),
\end{equation}
where $q(\basis_{a}) = p_a(\basis_{a}\,|\,\vecg_a,\,\M_a)$ is a normalized probability distribution over the states of $\basis_{a}\in\{\pm 1\}^{r_a}$. For an ICC $\M_a$, there is a bijective map between the $2^{r_a}-1$ parameters $\vecg_a$ of $\M_a$ and the $2^{r_a}-1$ state probabilities $q(\basis_{a})$ (the last probability being fixed by the normalization).
More precisely, the value of the parameters $g_{\vecmu}$ can be obtained from the states probabilities $q(\basis_{a})$ by taking the logarithm of \eqref{eq:ICC}, multiplying by $\phi^{\vecmu}(\spin)$ and summing over all the states of $\spin$. Finally using the orthogonality relation between operators \eqref{eq:completeness:1},  
one finds that for all interactions $\vecmu\in\M_a$:
\begin{align}\label{eqgb}
    g_{\vecmu}
    &= \frac{1}{2^n}\, \sum_{\spin}
    \phi^{\vecmu}(\spin)\,\log q\left(\basis_{a}(\spin)\right)\,.
\end{align}
Note that for $\vecmu\not\in\M_a$ this equation yields $g_{\vecmu}=0$, because the function $q\left(\basis_{a}(\spin)\right)$ is expressed solely in terms of the operators~$\vecmu\in\M_a$. 
In general, the statistical inference of ICCs (and therefore of MCMs) is more easily done in the $q$-representation, i.e. in terms of the parameters $q(\basis_{a})$. The result can then be re-written in the original spin model representation using~\eqref{eqgb}.\\ 

\paragraph*{\bf Maximum likelihood parameters.}
Using \eqref{eq:ICC}, the likelihood function of an ICC $\M_a$ for a dataset $\dataset$ can be written as: 
\begin{equation} \label{eq:likelihood}
P(\dataset\,|\,\vecg,\M_a)=\frac{1}{2^{N(n-r_a)}}\,\prod_{\basis_{a}}\,q(\basis_{a})^{k_{\basis_{a}}}\,,
\end{equation}
where the product is over all possible $2^{r_a}$ values of $\basis_a$, and where 
$k_{\basis_{a}}$ is the number of times the basis operators take the value ${\basis_a}$ over the entire dataset. Thanks to the bijection between the $2^{r_a}-1$ parameters $\vecg$ and the $2^{r_a}-1$ parameters $q(\basis_a)$, the maximum likelihood can be obtained by maximizing \eqref{eq:likelihood} directly over the parameters $q(\basis_a)$. 
This leads to the maximum likelihood estimate of $q$:
\begin{equation}\label{eq:max-likelihood:parameters}
\hat q({\basis_{a}})=\frac{k_{\basis_{a}}}{N}.
\end{equation}
Replacing these values in the likelihood function~\eqref{eq:likelihood} and combining with the factorization over ICCs in \eqref{factorization} leads to the value of the maximum likelihood of an MCM:
\begin{equation}\label{eq:max-likelihood}
P(\dataset\,|\,\vecghat,\,\mathcal{M})
    =\frac{1}{2^{N(n-r)}}\prod_{a\in\mathcal{A}}\left[
    \prod_{\basis_a}\,
    \left(\frac{k_{\basis_a}}{N}\right)^{\vspace{-4mm}k_{\basis_a}}\right]\,.
\end{equation}
Finally, Eq.~\eqref{eq:MCM_P_s:maxL} for the probability distribution of an MCM at the maximum-likelihood estimate is obtained by replacing the values of the maximum likelihood parameters~\eqref{eq:max-likelihood:parameters} in the factorization of \eqref{factorization} using that $p_a(\basis_{a}\,|\,\vecghat_a^{\prime},\,\M_a^{\prime}) = \hat{q}(\basis_{a})$.\\

\paragraph*{\bf Model evidence.} The calculation of the evidence of an ICC is likewise straightforward. We first exploit the invariance of Jeffreys prior under reparametrization~\cite{jeffreys1946invariant}, which takes the following form in the $q$-representation:
\begin{equation}
P_0(\boldsymbol{q}\,|\,\M_a)=\frac{\Gamma(2^{r_{a}-1})}{\pi^{2^{r_a-1}}}
    \prod_{\basis_{a}}\frac{1}{\sqrt{q(\basis_{a})}}
    \;\delta\Big(\sum_{\basis_{a}}q(\basis_{a})-1\Big),
\end{equation}
where we denote by $\boldsymbol{q}$ the vector of $2^{r_a}$ probabilities $q(\basis_{a})$ and $\Gamma$ is the gamma function. One can recognize the symmetric Dirichlet distribution with hyperparameter $\alpha=1/2$. 
Combined with the expression of the likelihood in~\eqref{eq:likelihood}, this yields the evidence of ICCs:
\begin{align}
P(\dataset\,|\,\M_a) & =\int\! d\vecg\; P(\dataset\,|\,\vecg,\,\M_a)\,P_0(\vecg\,|\,\M_a)
\label{app:eq:evidence:def}\\
 & =  \frac{1}{2^{N(n-r_{a})}}
 \frac{\Gamma(2^{r_{a}-1})}{\pi^{2^{r_a-1}}}
 \times\label{eq:multivariate_beta_fn}\\
 & \qquad \int_0^1 \hspace{-0.5mm}
 \prod_{\basis_{a}}\left[q(\basis_{a})^{k_{\basis_{a}}-\frac{1}{2}}\right]
 \delta\Big(\sum_{\basis_{a}}q(\basis_{a})-1\Big)
 \,d\boldsymbol{q}\nonumber\\
 & =\frac{1}{2^{N(n-r_{a})}}\,\frac{\Gamma(2^{r_a-1})}{\pi^{2^{r_a-1}}}\;\frac{\prod_{\basis_a}\Gamma(k_{\basis_a}+\frac{1}{2})}{\Gamma(N+2^{r_a-1})}\,,\nonumber
\end{align}
where one can recognize the multivariate beta function in the integral of \eqref{eq:multivariate_beta_fn}. 
This calculation easily generalizes to MCMs, in which the components $\M_a$ are associated with independent bases $\basis_a$.
Indeed, using~\eqref{factorization}, both the likelihood and the prior distribution $P_0$
factorize over the ICCs, which implies that the evidence in Eq.~\eqref{eq:evidence} is the product of the evidence of each ICC (reduced to their respective basis $\basis_a$).
This finally leads to the expression of the evidence of an MCM in Eq.~\eqref{eq:evidenceMCM} (we recall that $\Gamma(\frac{1}{2}) = \sqrt{\pi}$).

\section{Binary linear algebra formalism for spin models and gauge transformations}
\label{Materials:Linear_Algebra}

The following section gives more insight into the binary linear algebra formalism used in our codes. 
This framework also provides a better mathematical description of spin models and gauge transformations.
Consider an $n$-spin system $\spin = (s_1,\,\cdots,\,s_n)$ and the gauge transformation $\mathcal{T}_{\basis}:\;\spin\rightarrow \spin^\prime = \basis(\spin)$, 
defined by the set of $n$ independent operators $\basis(\spin) = (\phi^{\vecmu_1}(\spin),\, \cdots,\,\phi^{\vecmu_n}(\spin))$. As it is a bijection of the space of states, one can define the inverse transformation: $\spin = \basis^{-1}(\spin^\prime)$.\\

\paragraph*{\bf Matrix representation of gauge transformations.} 
We recall that the label $\vecmu$ indexing an operator $\phi^{\vecmu}$
is a binary vector encoding which spins contribute to the operator, such that 
$\phi^{\vecmu}(\spin) = \prod_{i=1}^n s_i^{\mu_i}$,
where $\mu_i\in\{0,1\}$ is the $i$-th element of $\vecmu$.
Similarly, we can write a gauge transformation $\mathcal{T}_{\basis}$ under the form of an $(n\times n)$-binary matrix:
\begin{align}\label{Methods:GT:Tb}
    \Tmat_{\basis} = (\vecmu_1,\, \cdots,\, \vecmu_{n})\,,
\end{align}
where the $n$ binary vectors $\vecmu_i$ defining the gauge transformation are now the columns of $\Tmat_{\basis}$.
The inverse gauge transformation $\basis^{-1}$ can then be obtained by
inverting the binary matrix $\Tmat_{\basis}$ over
$\mathbb{Z}/2\mathbb{Z}$, i.e. $\Tmat_{\basis^{-1}} = \Tmat_{\basis}^{-1}$ 
(when taking the matrix product over $\mathbb{Z}/2\mathbb{Z}$, the elements of the resulting matrix are taken modulo $2$).
Note that the fact that the set of operators $\{\phi^{\vecmu_1}(\spin),\, \cdots,\,\phi^{\vecmu_n}(\spin)\}$ is independent translates into the fact that the set of vectors $\{\vecmu_1,\,\cdots,\,\vecmu_n\}$ is linearly independent in $(\mathbb{Z}/2\mathbb{Z})^n$, and therefore that the rank of the matrix $\Tmat_{\basis}$ is $n$ and the matrix is invertable.\\

\paragraph*{\bf Binary representation of states and Transformed states.} Under the GT $\mathcal{T}_{\basis}$, a spin configuration $\spin$ of the system is transformed into the new configuration $\spin'=\basis(\spin)$. Introducing the binary variables $\alpha_i$ such that $s_i=(-1)^{\alpha_i}$ for all spins, 
the state~$\spin$ of the system is equivalently
represented by the binary vector $\boldsymbol{\alpha} = (\alpha_1, \,\cdots,\,\alpha_n)\in (\mathbb{Z}/2\mathbb{Z})^n$. 
Similarly, the transformed state $\spin'=\basis(\spin)$ is represented by the binary vector $\boldsymbol{\alpha}'$, whose elements $\alpha_i'$ are then given by the scalar product $\alpha_i' = \boldsymbol{\alpha}\cdot\vecmu_i$. 
Thus, using the matrix representation $\Tmat_{\basis}$ of gauge transformations in \eqref{Methods:GT:Tb}, the new state of the spin system 
is given by the vector-matrix multiplication over $\mathbb{Z}/2\mathbb{Z}$:
\begin{align}\label{Methods:GT:State}
    \boldsymbol{\alpha}' = \boldsymbol{\alpha}\; \Tmat_{\basis}\,, 
\end{align}
where $\boldsymbol{\alpha}$ and $\boldsymbol{\alpha}'$ are row vectors.
Similarly, a state in the original basis can be recovered from the state of the system in the new basis using:
\begin{align}\label{SM:eq:GT_inverse}
    \boldsymbol{\alpha} = \boldsymbol{\alpha}'\; \Tmat_{\basis}^{-1}\,. 
\end{align}

\paragraph*{\bf Transformed operator.} 
Any spin operator $\phi^{\vecmu}(\spin)$ can be transformed under a GT $\mathcal{T}_{\basis}$ 
by re-expressing $\phi^{\vecmu}(\spin)$ in terms of the new basis variables $\spin^\prime=\basis(\spin)$, as in Eq.~\eqref{App:eq:Phi_prime}. The transformed operator is then given by:
\begin{align}\label{Methods:GT:Operator}
    \mathcal{T}_{\basis}[\,\phi^{\vecmu}\,] = \phi^ {\vecmu^{\prime}}\,,
    \;\;\; {\rm where} \;\;\;\;
    \vecmu^{\prime} = \Tmat_{\basis}^{-1}\,\vecmu 
\end{align}
is a matrix-vector multiplication in $\mathbb{Z}/2\mathbb{Z}$, and $\vecmu$ and $\vecmu^\prime$ are two column vectors.\\

\paragraph*{\bf Transformed model.} The gauge transformation $\M^{\prime}$ of a model $\M$ is obtained by transforming each operator of $\M$, see Eq.~\eqref{App:eq:M_prime}. The probability of finding the system 
in the transformed state $\spin^\prime = \basis(\spin)$ in the transformed model $\M^\prime = \mathcal{T}_{\basis}[\M]$ is given by Eq.~\ref{GT:def_on_model}, where the vector of parameters $\vecg^\prime=\mathcal{T}_{\basis}[\vecg]$ is a permutation of the parameters $\vecg$ of $\M$ such that the transformed operator $\phi^{\vecmu'}$ in $\M'$ is parameterized by the same parameter as the operator $\phi^{\vecmu}$ in $\M$, i.e. (based on Eq.~\eqref{Methods:GT:Operator}):
\begin{align}\label{method:eq:g_prime}
    g_{\vecmu^{\prime}} = g_{\vecmu}
    \;\;\;\;{\rm for}\;\;\vecmu^{\prime} = \Tmat_{\basis}^{-1}\vecmu\,.
\end{align}

\paragraph*{\bf Example.} 
For $n=3$, take the gauge transformation $\spin^\prime=\basis(\spin) = (s_1s_2,\, s_2s_3,\, s_3)$. The inverse gauge transformation is $\spin=\basis^{-1}(\spin^\prime)=(s_1's_2's_3',\,s_2's_3',\,s_3')$. These two transformations can be written under the matrix forms:
\begin{align}
    \Tmat_{\basis} = 
    \begin{pmatrix}
        1 & 0 & 0\\
        1 & 1 & 0\\
        0 & 1 & 1
    \end{pmatrix}
    \qquad {\rm and} \qquad
    \Tmat_{\basis^{-1}} = 
    \begin{pmatrix}
        1 & 0 & 0\\
        1 & 1 & 0\\
        1 & 1 & 1
    \end{pmatrix}
\end{align}
One can easily check that $ \Tmat_{\basis}\; \Tmat_{\basis^{-1}} = \mathbb{I}_3$.\\

Take the state $\spin = (-1,\,-1,\,+1)$, which can also be defined as $\boldsymbol{\alpha}=(1,\,1,\,0)$ using that $s_i=(-1)^{\alpha_i}$. 
In the new basis $\basis(\spin)$ defined above, this state takes the new value $\spin' = (+1,\,-1,\,+1)$, which can also be written as $\boldsymbol{\alpha}'=(0,\,1,\,0)$. One can check that the same value of $\boldsymbol{\alpha}'$ can be obtained using $\boldsymbol{\alpha}' = \boldsymbol{\alpha}\; T_{\basis}$, as in \eqref{Methods:GT:State}.\\

Take the operator $\phi^{\vecmu} (\spin)=s_1s_3$, which is identified by the binary vector $\vecmu = (1,\,0,\,1)$. One can write the transformed operator $\mathcal{T}_{\basis}[\phi^{\vecmu}]$ using the inverse transformation $\spin =\basis^{-1}(\spin')$. In particular, using that $s_1 = s_1's_2's_3'$ and that $s_3 = s_3'$, one gets:
\begin{align}
    \phi^{\vecmu'}(\spin')=(s_1's_2's_3')\times(s_3') = s_1's_2'\,,
\end{align}
which is identified by the binary vector $\vecmu' = (1,\,1,\,0)$. One can check that the same value of $\vecmu'$ is obtained by using 
$\vecmu' = \Tmat_{\basis}^{-1}\,\vecmu$ as in \eqref{Methods:GT:Operator}.

\section{Search for the best IM}\label{Materials:proofIM}

An independent model (IM) is an MCM for which all the components $\M_a$ only contain a single operator (i.e., $r_a=1$ for all $a\in\mathcal{A}$). 
Let us focus on the case where the number of components $|\mathcal{A}|$ is equal to the number of spins $n$.  
All such IM have $n$ independent operators, $\M = \{\phi_1(\spin),\,\cdots,\,\phi_n(\spin)\}$. 
Model selection among these models only requires comparing their likelihood, because they all have the exact same complexity (as they belong to the same complexity class~\cite{beretta2018stochastic}).
The maximum log-likelihood of an independent model takes the simple form:
\begin{align}\label{eq:logL_IM}
    \log P(\dataset\,|\,\boldsymbol{\hat{g}},\M)=-N\sum_{a=1}^n H[\phi_a(\spin)]\,,
\end{align}
where $H[\phi_a(\spin)]$ is the entropy of the binary operator $\phi_a(\spin)$ in the dataset $\dataset$:
\begin{align}
    H[\phi_a] = - p_a \log p_a - (1-p_a) \log(1-p_a)\,,
\end{align}
where $p_a=\mathbb{P}[\phi_a(\spin)=1]$ is the empirical probability that $\phi_a(\spin)=1$ in the dataset. We note that the entropy of an operator can equivalently be written in terms of the bias $m_a=p_a-(1-p_a)$
of the operator in the data:
\begin{align}
    &H[\phi_a] = -\frac{1+m_a}{2}\log\left(\frac{1+m_a}{2}\right)-\frac{1-m_a}{2}\log\left(\frac{1-m_a}{2}\right) \nonumber\\
    &{\rm where}\;\;
    m_a = \frac{1}{N}\sum_{i=1}^N  \phi_a\left(\spin^{(i)}\right) = \langle \phi_a(\spin) \rangle_D\,.
\end{align}
The entropy $H[\phi_a]$ is maximal when $p_a=0.5$ (or equivalently when the operator is least biased, $m_a=0$) and is minimal  
when $p=0$ or $p=1$ (or equivalently, when the bias of the operator is maximal, $|m_a|=1$). 
\eqref{eq:logL_IM} thus implies that the most likely independent model is given by the most biased (or least entropic) set of $n$ independent operators.

The exhaustive algorithm to find the most biased set of independent operators among a pool of possible operators is the following. First, compute the bias of all the operators and rank them from the most to the least biased. This procedure is of the order $O(\bar{N}K\log K)$, where $\bar{N}$ is the number of different states observed in the dataset, and $K$ is the total number of operators to compare (if one considers all possible operators, then $K=2^n-1$). This is the most computationally expensive part of this algorithm. Next, we must extract the most biased independent set. To do so, we create an $(n \times K)$-binary matrix in which each column contains the binary representation of one of the $K$ operators (we recall that the vector $\vecmu_i$ is the binary representation of the operator $\phi^{\vecmu_i}$). 
Operators are placed by decreasing order of their bias, starting with the most biased operator in the first column of the matrix.
This approach is similar to the way we defined the binary matrix representation of a gauge transformation in Eq.~\eqref{Methods:GT:Tb}, 
except that now the columns are not necessarily independent from each other. To identify the most biased independent operators, we then use Gaussian elimination: the columns with leading coefficients will finally identify the most biased independent operators.

In the main text, we propose a heuristic iterative algorithm to find a good basis when the dimension of the system is too large. All the steps above are then still valid, the only difference is the initial pool of operators from which we want to identify the best basis. 
In the case of the exhaustive search, this pool  
contains all the $2^n-1$ operators of the $n$-spin system; while in the heuristic algorithm, it contains only the operators up to a fixed order $k$ (in a given representation).

When $K$ is large, there are other ways to accelerate the algorithm slightly. 
Ideally, one would like to reduce the number of operators to rank (as this is computationally demanding). This can be done by excluding operators whose bias is too low, for instance 
by using a threshold on the $p$-value for the bias of the selected operators. 
Note that this can lead to a set of independent operators that is smaller than $n$, meaning that some dimensions would then be considered irrelevant.
Another straightforward way of accelerating the algorithm when $K$ is large is to reduce the size of the matrix to contain only a fixed number of operators (for instance $100$ columns). One can then perform the Gaussian elimination several times, by keeping leading columns and adding new operators as needed, 
until $n$ independent operators are found.

\section{Sampling from an MCM}\label{Materials:Sampling} 
Consider the MCM $\M=\cup_{a\in\mathcal{A}} \M_a$ of an $n$ spin system~$\spin$
and a preferred basis $\basis(\spin) = \cup_{a\in\mathcal{A}} \basis_a(\spin)$ of $\M$. 
We denote $r=\sum_{a\in\mathcal{A}}r_a$ the rank of $\M$.
For a given dataset, the probability distribution of 
$\M$ at best fit reads:
\begin{equation}
\label{eqMCMmaxlik}
P(\spin\,|\,\boldsymbol{\hat{g}},\M)=\frac{1}{2^{n-r}} 
\prod_{a\in\mathcal{A}}\frac{k_{\basis_a(\spin)}}{N},
\end{equation}
where $k_{\basis_a}/N = \hat{q}({\basis_{a}})$ are the empirical probabilities of observing the states $\basis_a(\spin)$ in the data. 
We recall that each distribution $\hat q({\basis_{a}})$ is normalized over the $2^{r_a}$ states of $\basis_a$ (see \eqref{eq:ICC}). To sample a configuration $\spin$ from this model, we perform the following steps:
\begin{itemize}
    \setlength{\parskip}{0pt}
    \item [1)] For each ICC $\M_a$, we draw a random configuration of $\basis_a$ with probability $\hat{q}(\basis_a)=k_{\basis_a}/N$.
This can be done either by sampling directly from the empirical distribution $\hat{q}(\basis_a)$, or by drawing uniformly a configuration $\spin^{(i)}$ from the data and then computing the transformation $\basis_a(\spin^{(i)})$. Eq.~\eqref{eqMCMmaxlik} ensures that this procedure samples a value of $\basis_a$
with the correct distribution. 
    \item [2)] If $r<n$, then one must complete $\basis(\spin)$ with $(n-r)$ independent operators,
$\bar{\basis}(\spin)=(\phi_{r+1}(\spin),\,\cdots,\,\phi_n(\spin))$, to form a basis of the whole $n$-dimensional space.
The resulting basis $\boldsymbol{B}(\spin)=\basis(\spin)\cup\bar{\basis}(\spin)$ defines a gauge transformation $\mathcal{T}_B$ 
between the original data states~$\spin$ and the new states $\spin^\prime = \boldsymbol{B}(\spin)$.  
The states of the $(n-r)$ spins not modeled by $\M$ are then randomly sampled with probability $1/2$ to be $+1$ or $-1$, 
which produces a sampled value of $\bar{\basis}$.
Note that for all the applications discussed in this paper, we searched for a best basis $\basis^*$ of the whole $n$-spin system (and then for an MCM on this basis). Therefore, for all these cases, $r=n$ and we didn't need to introduce a complementary basis $\bar{\basis}$. 
    \item [3)] Next, we concatenate the sampled states $\basis_a$ and $\bar{\basis}$ to obtain a sampled value of $\spin^\prime=(\cup_{a\in\mathcal{A}}\basis_a)\cup\bar{\basis}$.
    \item [4)] Finally, we perform an inverse GT to obtain a sampled value of 
$\spin=\boldsymbol{B}^{-1}(\spin^\prime)$. 
As described in Sec.~\ref{Materials:Linear_Algebra}, 
this last step requires the inversion of an $n\times n$ binary matrix modulo~2 (see \eqref{SM:eq:GT_inverse}). 
If one wants to sample multiple states from the same MCM, then this inversion only needs to be performed once. 
\end{itemize} 

Our implementation of this inversion (using Gaussian elimination) and of the sampling functions can be found in Ref.~\cite{Greedy_Algo}.
In practice it is computationally very efficient to sample from an MCM, as it only requires drawing $|\mathcal{A}|$ independent configurations from the dataset
and one binary vector-matrix multiplication.
Step 1) exploits the fact each ICC $\M_a$ is complete on the variables $\basis_a$, and therefore, computing $\basis_a(\spin)$ on a randomly drawn datapoint $\spin$ from the dataset $\dataset$ directly generates values of $\basis_a$ that are distributed according to~\eqref{eqMCMmaxlik}. 
Hence, a new state of $\basis$ can be generated in a straightforward manner from $|\mathcal{A}|$ independent draws from the dataset.

\section{Enumeration}
\label{app:enum}
\paragraph*{\bf Number of Independent Models.}
All the IMs with $r$ interactions can be obtained by gauge transformations of the IM $\M_{ind}=\{s_1,\ldots,s_r\}$. Therefore, their number is equal to the number of GTs that transform this model into different ones and is given by:
\begin{equation}
\label{eq:IMcardinality}
    \mathcal{N}_{ind}(n,r) = 
    \frac{1}{r!} \prod_{i=0}^{r-1} \left(2^n-2^i\right).
\end{equation}
This corresponds to the number of ways of choosing $r$ independent operators in the set of all possible $2^n-1$ operators, divided by the number of possible permutations of these $r$ elements. The first operator can be chosen in $2^n-1$ ways. Then, after having chosen $i$ independent operators, the $(i+1)$-th 
operator can be chosen among $2^n-2^i$ operators, because $2^i$ operators are now dependent on the first $i$ operators already selected. 
For large values of $n$, the number of IMs of rank $r$ grows as $2^{n r}/r!\,$. 
Finally, the total number of IMs in an $n$-spin system is obtained by summing \eqref{eq:IMcardinality} over all possible values of $r$ (from $r=0$ to $n$) and is displayed in Fig.~\ref{Fig2:NbModels} (labeled as ``IM''). We estimated that this number grows roughly as $2^{n^2}$.\\

\paragraph*{\bf Number of Pairwise Models.} Since there are $n(n+1)/2$ possible single-spin and pairwise interactions, there are in total $2^{n(n+1)/2}$ possible pairwise models (see plot labeled ``PM$^*$'' in Fig.~\ref{Fig2:NbModels}). This number grows slower than the number of Independent Models.\\

\paragraph*{\bf Number of Sub-Complete Models.}
In an $n$-spin system, all the SCMs of a given rank $r$ can be obtained by taking all possible GTs of any chosen SCM $\M$ of rank~$r$. As the GTs that involve only the operators in $\M$ leave $\M$ invariant, 
the number of distinct SCMs is therefore: 
\begin{align}\label{eq:SubCompleteM:Class:nbOfM}
    \mathcal{N}_{\rm ICC}(n,r) = \prod_{i=0}^{r-1}\frac{2^n - 2^i}{2^r - 2^i}\,.
\end{align}
Here the numerator $ \prod_{i=0}^{r-1}(2^n - 2^i)$ counts the number of ways of choosing the $r$ basis operators of the SCM.
The denominator $ \prod_{i=0}^{r-1}(2^r - 2^i)$ counts the number of GTs that transform the basis of the SCM while leaving $\M$ invariant. Finally, the total number of SCMs (i.e of models with a single ICC) in an $n$-spin system is obtained by summing \eqref{eq:SubCompleteM:Class:nbOfM} over all possible values of $r$ (from $r=0$ to $n$) and is displayed in Fig.~\ref{Fig2:NbModels} (labeled as ``SCM'').\\

\paragraph*{\bf Number of Minimally Complex Models.}
Consider an MCM with $m$ ICCs for an $n$-spin system; we denote $m_{r_a}$ the number of ICCs of rank $r_a$, such that $m = \sum_{r_a=0}^n m_{r_a}$.
For MCMs on $n$ spins with $m_{r_a}$ ICCs of the same rank ${r_a}$ 
one can combine the two previous arguments and get the number of such distinct models:
\begin{align}
    \mathcal{N}_{\rm MCM}(n,\{m_{r_a}\}) = 
    \frac{\displaystyle\prod_{i=0}^{r-1}\left(2^n-2^i\right)}
    {\displaystyle\prod_{r_a=1}^n m_{r_a}!\,\left[\prod_{i=0}^{r_a-1}(2^{r_a}-2^i)\right]^{m_{r_a}}}\nonumber
\end{align}
where the factor $1/m_{r_a}!$ accounts for permutations among ICCs of the same size. To count the total number of MCMs with $n$ spins (plotted in Fig.~\ref{Fig2:NbModels}), one must sum over all the classes of MCMs with different degeneracies $m_{r_a}$. The number of classes with different degeneracy corresponds to the number of integer partitions of the rank $r$ of the MCM in the form $r=\sum_{r_a} {r_a} m_{r_a}$, and summing over all $r\le n$. For instance, $r=8$ admits $22$ partitions which are:
\begin{gather}
    8, 71, 62, 611, 53, 521, 5111, 44, 431, 422, 4211,\nonumber\\
    41111, 332, 3311, 3221, 32111, 311111, 2222, \nonumber\\
    22211, 221111, 2111111, 11111111\,.\nonumber
\end{gather}
Here, the partition denoted ``$8$'' corresponds to the class of MCMs that have only one ICC (with rank $r=8$). 
The partition ``$422$'' is the class of  MCMs formed of three ICCs, 
two of rank 2 and one of rank 4.\\

\paragraph*{\bf Number of Minimally Complex Models with the same preferred basis.}
For a given choice of $n$ basis operators, the number of MCMs of rank $r=n$ that admits that basis as a preferred basis is given by the number of possible partitions of this set of $n$ basis elements, which is 
the Bell number $B_n$. Note that Ref.~\cite{berend2010improved} gives the upper-bound 
$B_n<\big(\frac{0.792\,n}{\ln(n+1)}\big)^n$. 
Finally, the total number of MCMs that share the same preferred basis (indicated as ``MCM$^*$'' in Fig.~\ref{Fig2:NbModels}) is obtained by summing this number over all ranks $r$ (from $r=0$ to $n$) while accounting for the possible choices of the $r$ basis operators:
\begin{align}
    \mathcal{N}_{MCM^{*}}(n) = \sum_{r=0}^{n} \binom{n}{r} B_r\,.
\end{align}
For example, for $n=9$, 
there are $115\,975$ MCM$^*$ in total (which are the MCMs that share the same preferred basis); only $21\,147$ of them have the rank $r=9$.

\section{Toy example: Recovering the boolean description of a non-noisy binary dataset}
\label{app:ToyModel}

\begin{figure}[h!]
    \centering
    \includegraphics[width=\linewidth]{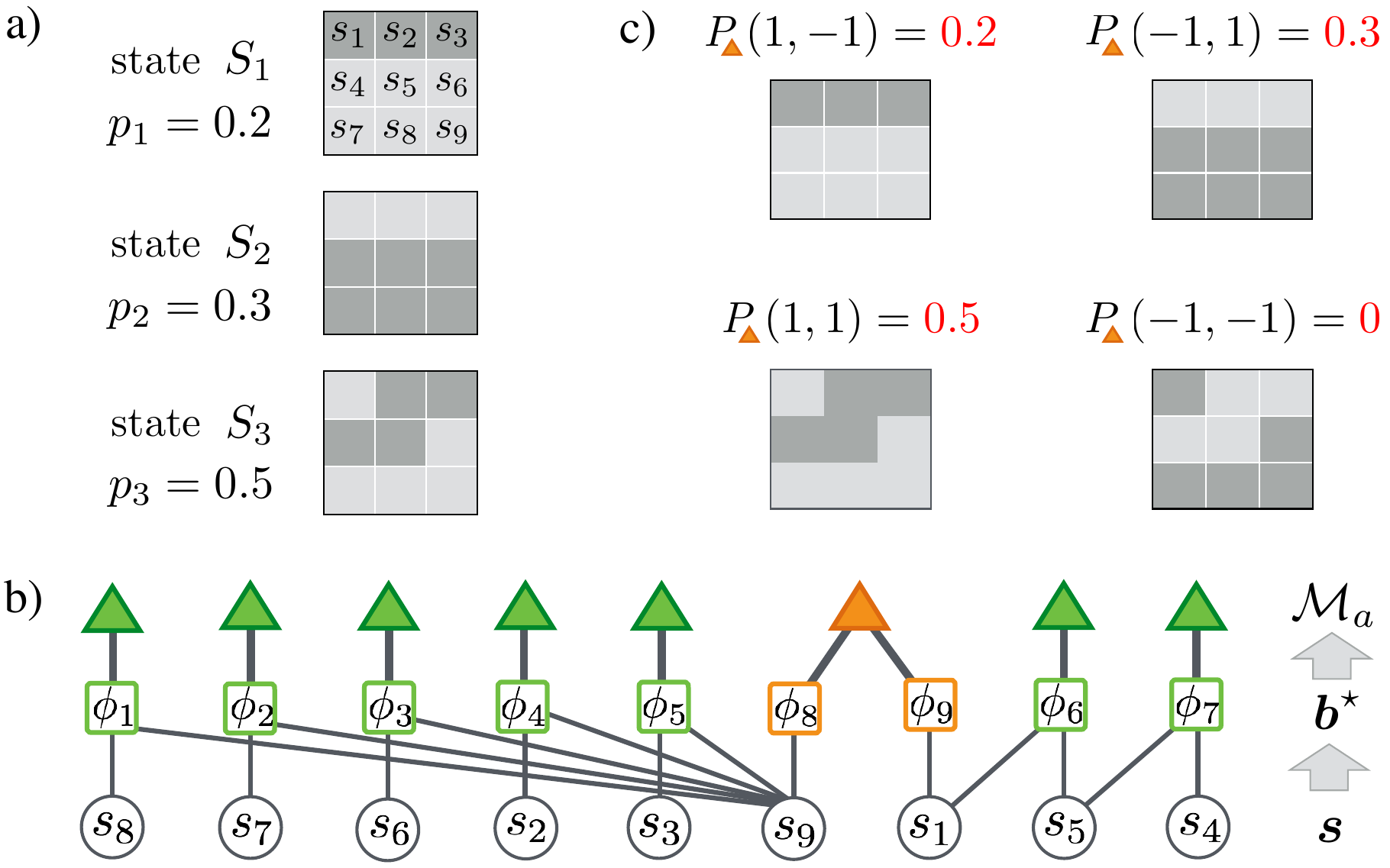}
    \caption{{\bf Analysis of a simple artificial dataset.} {\bf a)} We generated a dataset composed of three different states ($S_1$, $S_2$ and $S_3$) each sampled with a different probability (respectively $p_1$, $p_2$ and $p_3$). Dark pixels correspond to the spin state $s_i=-1$, and light pixels to $s_i=+1$. {\bf b)} Factor graph of the best MCM found for this dataset. {\bf c)} Once it is fitted, the best MCM can generate four different states. Three of them are the original states and are sampled with the same probability than their occurrence in the original dataset, the last one is sampled with probability $0$.
}
\label{Fig5:shapes}
\end{figure}
We tested our approach on an artificial dataset of $n=9$ variables and $N=10^5$ datapoints that we generated by sampling randomly three different states, $S_1$, $S_2$, and $S_3$, with respective probabilities $p_1=0.2$, $p_2=0.3$, and $p_3=0.5$ (see Fig.~\ref{Fig5:shapes}.a).
There were no additional patterns or sources of noise in the data.
The size of the system ($n=9$) was chosen sufficiently small to easily perform the exhaustive search described in the previous section. The factor graph of the best MCM found for this dataset is shown in Fig.~\ref{Fig5:shapes}.b. It is composed of $8$ ICCs, $7$ of which are based on a single operator (green triangles).

Once it is fitted, this MCM can be used as a generative model.
Notably, fitting an MCM does not require inferring the vector $\vecg$ of parameters, thanks to the bijection between the parameters $\vecg_{a}$ and the state probabilities $p(\boldsymbol{b}_a)$ for each ICC.
Instead, the likelihood is simply maximized when, for all the states of $\basis_a$, the probability of occurrence of the state in the model is equal to its probability of occurrence in the data (see~\eqref{eq:max-likelihood:parameters}). 
We thus obtain that all the single-operator ICCs in Fig.~\ref{Fig5:shapes}.b (green triangles) are associated to statements that are true with probability $1$ over the entire dataset, i.e. they are associated to independent tautologies of the data.
For instance, in this dataset, the operator $\phi_1 = s_8s_9$ is equal to $1$ with probability $1$ and the operator $\phi_5 = s_3s_9$ is equal to $-1$ with probability $1$, i.e., the two statements $\phi_1 =1$ and $\phi_5 =-1$ are true over the entire dataset.
After taking into account the $7$ independent tautologies, we are left with only four states that could be generated by the best MCM (see Fig.~\ref{Fig5:shapes}.c), each of them corresponding to a different value of the basis elements $\basis_a=(\phi_8, \phi_9)$ of the last ICC (orange triangle). 
Fitting the probabilities of occurrence of $\basis_a$ on the data, we obtained that $P(\phi_8=1,\phi_9=-1)=0.2$, $P(-1,1)=0.3$, $P(1,1)=0.5$ and, by normalization, $P(-1,-1)=0$ (there are only three independent probabilities). This finally provides a full description of the rules used to generate the data.
In this case, we observe that the best MCM provides the shortest complete description of the data in terms of spin models.
Note that because each spin operator is associated to an elementary boolean function~\footnote{A spin state can be re-written as $s_i=(-1)^{x_i}$ where $x_i\in\{0,1\}$. Thus the spin operator $\phi^{\vecmu}(\spin)=\prod_{i\in\vecmu}s_i$ corresponds to the boolean function $F^{\vecmu}(\x)=\bigoplus_{i\in\vecmu} x_i$, where $\oplus$ is the XOR operator, and $\phi^{\vecmu}(\spin)=(-1)^{F^{\vecmu}(\x)}$.}, the best MCM therefore also provides a complete decomposition of the data in terms of such boolean functions,
each associated with a parameter that is the probability that this function takes value $1$ over the dataset ($7$ of which being just probability $1$ or $0$).\\

\section{Complementary Figures}\label{Materials:Figures}

\begin{figure}[h!]
\centering
\includegraphics[width=\linewidth]{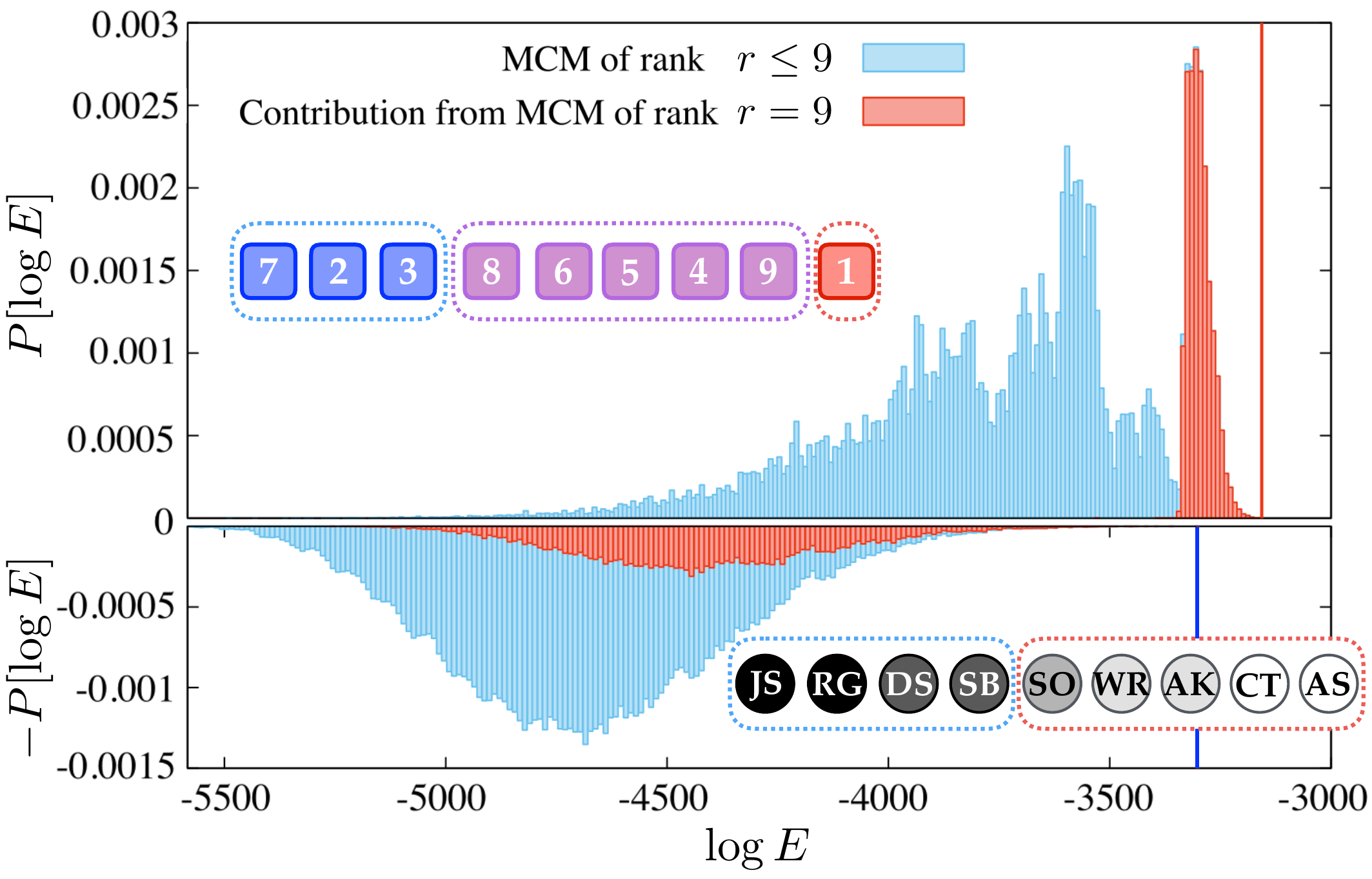}
\caption{(colors online)
    {\bf Distribution of the log-evidence of MCMs for the US Supreme court data}: for all the MCMs that admit the best IM $\basis^*$ as a preferred basis (blue distribution in the top graph) and for all the MCMs that admit the original basis of the data as a preferred basis (blue distribution in the bottom graph). {\bf Top plot.} The best MCM overall is based on $\basis^*$ (see Fig.~\ref{Fig4:SCOTUS}.b); it has three ICCs
    and a log-evidence of $\log E \simeq -3327$ (indicated by the red vertical line). 
    The basis elements of $\basis^*$ are represented by squares, numbered as in Fig.~\ref{Fig4:SCOTUS}.b, and the three ICCs are indicated by different colors (blue, purple, red).
    {\bf Bottom plot.}
    The best MCM based on the original basis of the justices is shown here; 
    it has two ICCs, indicated by dotted squares (blue or red), and has a log-evidence of $\log E\simeq -5258$ (indicated by the blue vertical line). 
    {\bf In both graphs.}
    The distribution in blue corresponds to the distribution for all the MCMs 
    with the same preferred basis (i.e., models of all rank from $r=0$ to $r=9$).
    With $n=9$ spins, there are $115\,975$ such models. 
    A part of this distribution has been highlighted in red; it corresponds to the subset of MCMs with rank $r=9$. With $n=9$ spin, there are $21147$ such models. It is interesting to observe that most MCMs based on the original basis (blue distribution in the bottom graph) have a significantly lower log-evidence than the MCMs of rank $r=9$ based on the best IM $\basis^*$ (red distribution in the top graph).
    }
\label{fig:Distrib_log_evidence}
\end{figure}

\begin{figure}[h!]
\centering
\includegraphics[width=\linewidth]{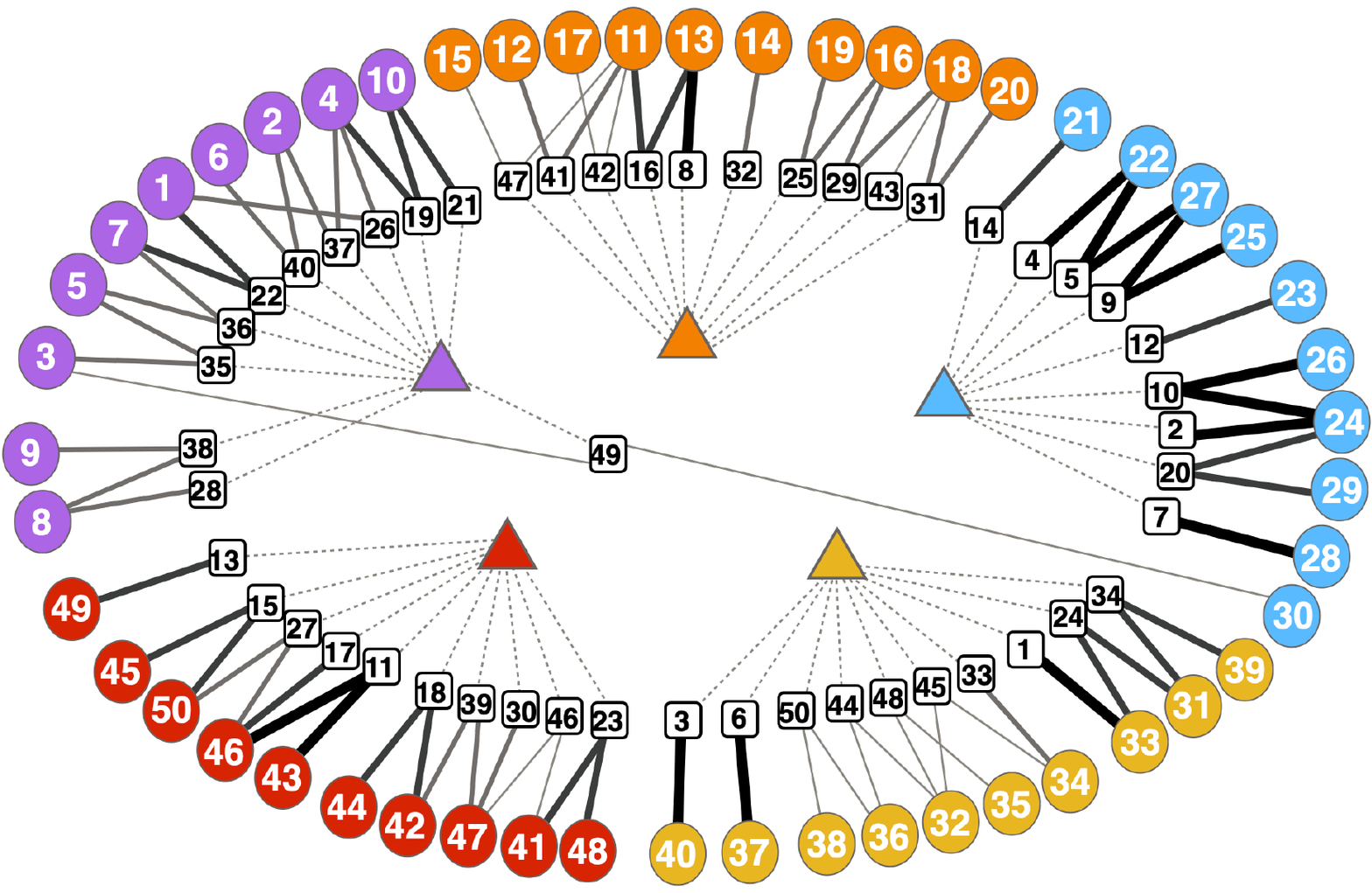}
\caption{(colors online)
    {\bf Factor graph representation of the best MCM found for the Big 5 dataset:} Original variables are represented by circles numbered from 1 to 50 (for the 50 statements, ordered in the same order as the dataset of Ref.~\cite{Big5data}).
    The best basis was found with the iterative search procedure at order $k=4$ (which takes about 3 hours on a laptop). Basis operators are numbered from the strongest to the weakest and the strength of the interactions is also indicated by the thickness and opacity of the link (darker-thicker links indicate stronger interactions). 
    We observe several strong dependencies between questions within the Agreeableness ICC (in blue); in particular, the correlations between questions 22 ({\it I'm interested in people}) and 27 ({\it I am not really interested in others}), 
    27 and 25 ({\it I am not interested in other people's problems}), 
    and between 24 ({\it I sympathize with other people's problems.}) and 26 ({\it I have a soft heart}) are the strongest in the system. Such information provided by the best basis could be used to improve the design of the questions.
    The ICCs of the best MCM$^*$ found in this basis are indicated by the triangles. This is the same MCM as the one displayed in the original basis in Fig.~\ref{Fig5:Big5}.
    }
\label{fig:big5:Best_MCM}
\end{figure}

\section{Comments on our heuristic algorithms}
\label{app:HeuristicAlgo:comments}

\paragraph*{\bf Greedy search for the Best MCM.}
For the Big~5 dataset, the Greedy search was only able to find the best MCM (displayed in Fig.~\ref{fig:big5:Best_MCM} and~\ref{Fig5:Big5}, with $\log E = -35.80$ bits/datapoint) from the search in the original basis, but wasn't able to find it in the new basis. Instead, it found a sub-optimal MCM with $\log E = -36.12$ bits/datapoint. Our understanding is that, because the greedy algorithm starts by making local merging steps, it is more likely to initially take a wrong path in the new basis (where basis variables are the most independent) than in the original basis (where correlations between variables are much stronger). 
Yet, for most of the datasets we analyzed with the Greedy search algorithm, 
we found a more optimal MCM in the best basis than in the original basis. 
In the case of the Big~5, the correlations between the questions are, by design, mostly prevalent within the traits rather than between the traits, 
which means that the original basis of the data is likely a preferred basis for the best MCM.
This in part explains why it was possible to find a better MCM with the greedy search in the original basis than in the new basis.

For the MNIST data, the Greedy search found a more optimal MCM in the best basis than in the original basis (see Fig.~\ref{fig:MNIST:SI}.c). 
In particular, the average ICC size is smaller in the best basis ($\overline{r}_a\simeq 7.12$) than in the original basis ($\overline{r}_a\simeq 8.07$),
resulting in a large reduction of the number of interactions $K_{MCM_{best}}=34\,631$ in the best basis compared to $K_{MCM_{0}}=73\,251$ in the original basis (see Fig.~\ref{fig:MNIST:SI}.d).
This is a large number of parameters, but it is still 
smaller than the number of different states observed in the dataset (which is $57\,903$). We believe that this MCM is likely not the most optimal one,  
but it is still able to sample states that resemble vaguely digits (see Fig.~\ref{Fig6:MNIST}).
In the future, it would be useful to develop more efficient algorithms to solve this optimization problem.

\paragraph*{\bf Heuristic search for the Best Basis.}
Fig.~\ref{fig:MNIST:SI}.b shows the number of basis operators of order $k$ obtained for different values of $k_{max}$ by the heuristic ``Best Basis'' search described in Sec.~\ref{Sec:LargeSystems} for the MNIST dataset.  
We observe that the distribution doesn’t change much from $k_{max}=2$ to $k_{max}=4$. This indicates that, although the number of variables is reasonably large ($121$), the algorithm was able to explore most of the relevant high-order basis operators, even with small values of $k_{max}$. 

\begin{figure*}[h!]
\centering
\includegraphics[width=0.95\linewidth]{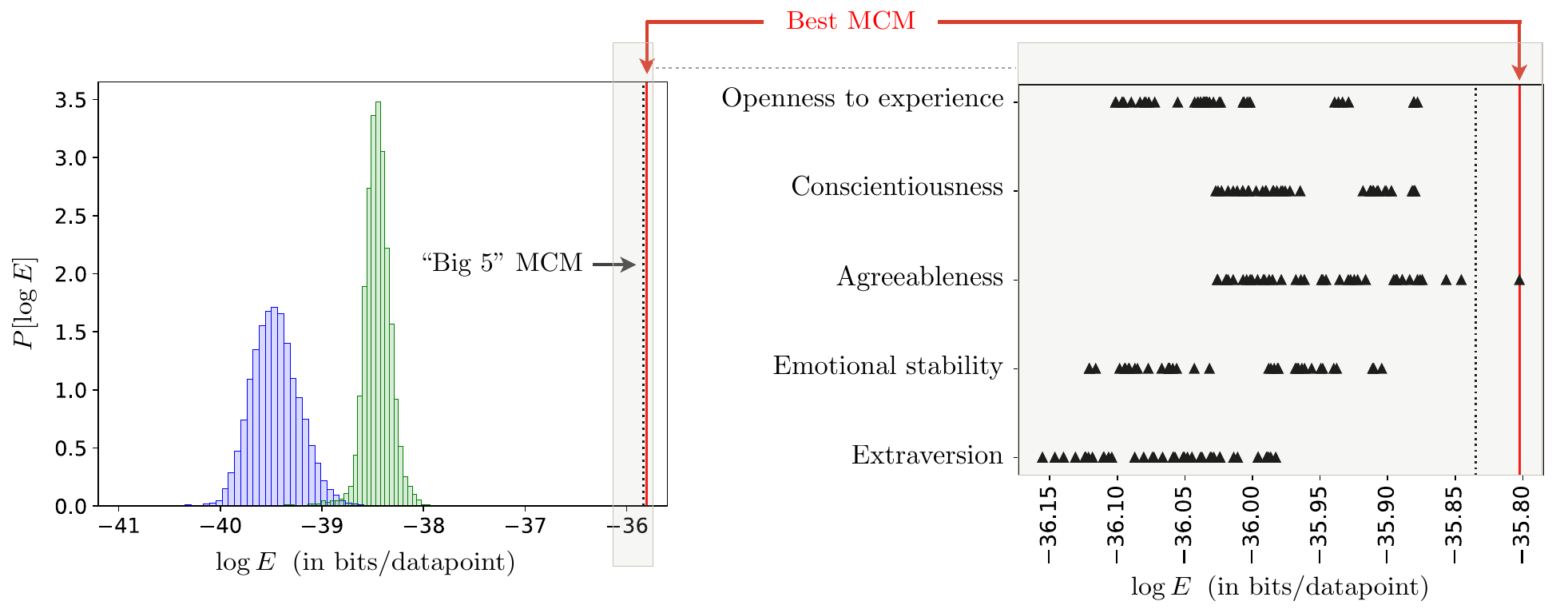}
\caption{(colors online)
    {\bf Left. \,Big 5 dataset: Distribution of the log-evidence of randomly sampled MCMs with 5 ICCs.} 
    We uniformly sampled $10^4$ partitions of the $50$ variables into 5 ICCs, both in the original basis (blue distribution) and in the best basis (green distribution).
    The log-evidence of the best MCM (see Fig.~\ref{fig:big5:Best_MCM}) is indicated by a red vertical line ($\log E \simeq -35.802$)
    and is just slightly larger than the log-evidence of the MCM partitioning the questions into the five standard categories of the Big 5 personality test,
    indicated by a vertical dashed line ($\log E \simeq -35.835$).
%
    Both values are far in the right tail of the two distributions, a good indication that both models capture interesting features of the data.
    Among all the $2.10^4$ randomly sampled partitions,
    the largest log-evidence found is $\log E\simeq -37.89$ bits/datapoint
    and the lowest is $\log E\simeq -42.90$ bits/datapoint (outside of the window). 
    Note that there are $\binom{50}{5}= 2\,118\,760$ possible partitions of the $50$ variables into $5$ parts.
    {\bf Right. \,Variation of the log-evidence of the ``Big 5 MCM'' when moving one question to another ICC:}
    This figure displays the value of the log-evidence of the corresponding MCM when moving one question from its original ``Big 5 ICC'' 
    (see labels on $y$-axis) to another ICC.
    For instance, 
    moving question 30 from ``agreeableness`` to the ``extroversion'' ICC (see right-most triangle on top of the vertical red line)
    increases the log-evidence compared to the MCM corresponding to the standard Big 5 partition (indicated by the dashed line) and gives the Best MCM (red line). Moving any other question will only decrease the evidence, which is a good indication that the questions 
    are most indicative of the traits they are meant to assess.
    Based on the data,
    the ``agreeableness`` ICC appears to be the least distinguishable from the other traits. 
    For example, the two next right-most triangles in the ``agreeableness`` ICC correspond to moving question Q.23 (``I insult people'') to the ``emotional stability'' ICC
    ($\log E\simeq -35.845$) or to the ``conscientiousness'' ICC
    ($\log E\simeq -35.857$).
    The fact that these moves don't seem to decrease the log-evidence significantly indicates that this question is also informative of these two other traits.
    On the contrary, the questions probing Extraversion are mostly informative of that specific trait.
    }
\label{fig:big5:Distrib_log_evidence}
\end{figure*}

\begin{figure*}[h!]
\centering
    \includegraphics[width=0.8\linewidth]{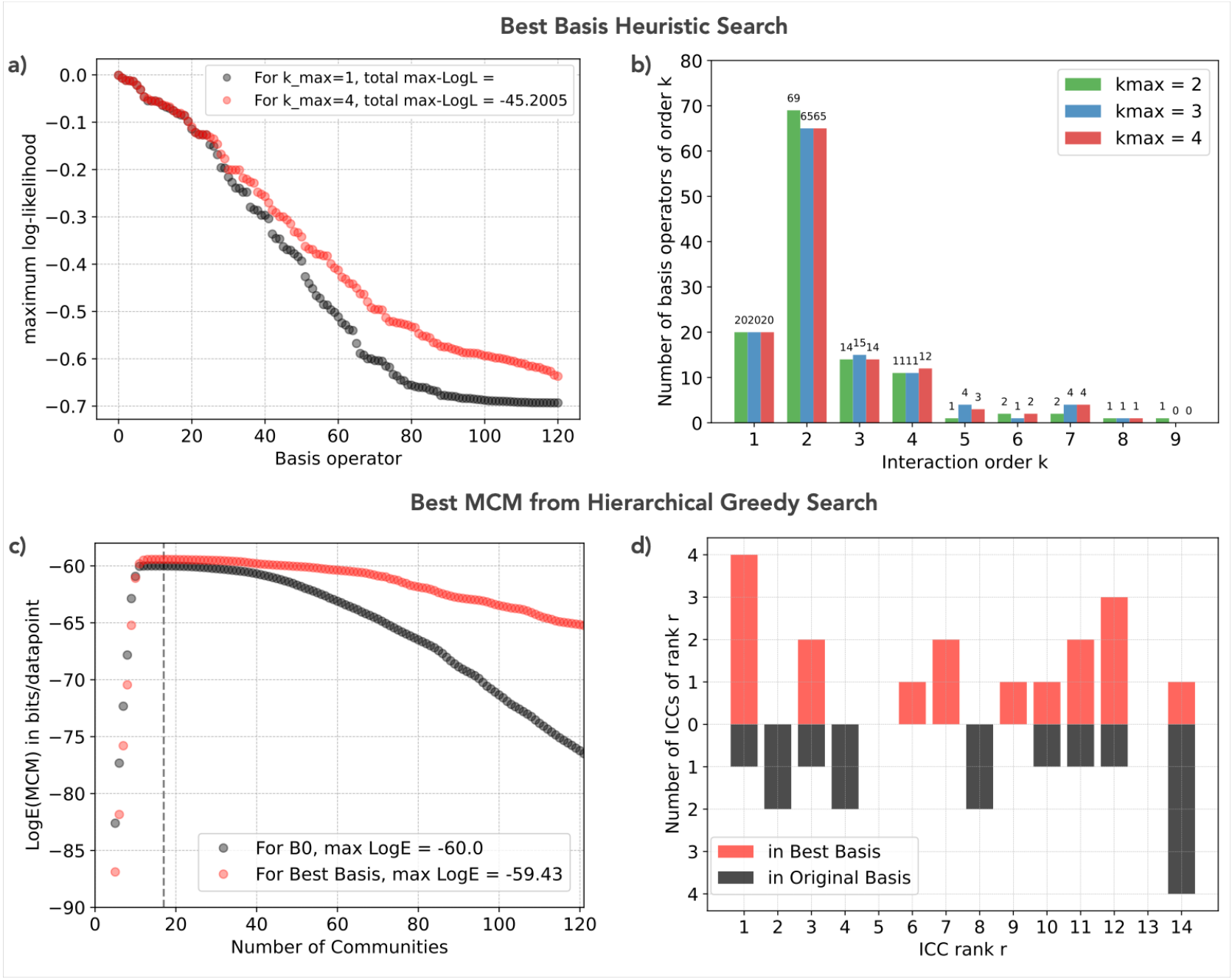}
\caption{(colors online)
    {\bf a)} Maximum likelihood of the basis operators, ordered from the largest (i.e. most biased operator) to the lowest, for the original basis (in black) and for the best basis found at $k_{max}=4$ (in red). By definition (see App.~\ref{Materials:proofIM}), more biased operators have a larger maximum likelihood. The most biased basis operators are mostly identical between these two bases (overlap of the curves on the left part) and include the $20$ single-spin operators appearing in panel b). 
    The least biased operators, in the right tails of the plots, are different when going from the original basis to the best basis (their maximum likelihood is larger in the best basis, as expected).
    {\bf b)} Number of basis operators of order $k$ obtained for different values of $k_{max}$ for the heuristic Best Basis search described in Sec.~\ref{Sec:LargeSystems}. At $k_{max} = 1$, all $121$ basis operators have order $k=1$ (i.e. all field interactions). We observe that the distribution doesn’t change much from $k_{max}=2$ to $k_{max}=4$. This indicates that the algorithm is able to explore most of the relevant high-order basis operators, even at relatively small values of $k_{max}$.
    {\bf c)} Evolution of the log-evidence of the MCMs along the greedy merging procedure performed in the original basis (in black) and in the new basis (in red) as the number of ICCs goes from $A=121$ (independent model) to $A=1$ (complete model). The maximum is reached a little sooner in the new basis (at $A=17$) than in the original basis (at $A=15$), which means that the best MCM found in the new basis has on average smaller communities. Finally, the algorithm reaches a maximum in log-evidence that is 
    larger ($\log E = -59.43$ bits/datapoint) in the best basis than in the original basis ($\log E = -60.0$ bits/datapoint).
    {\bf d)}~Distribution of the number of ICCs of a given rank $r_a$ for the best MCM found by the Greedy algorithm in the original basis (in black) and in the best basis (in red). We observe that in the new basis, the MCM has ICCs of lower rank than in the original basis (e.g., more ICCs of rank $r_a=1$ and fewer ICCs of rank $r_a=14$), 
    resulting in a large reduction of the number of interactions in the best basis ($K_{MCM_{best}}=34\,631$) compared to the original basis ($K_{MCM_{0}}=73\,251$). 
    }
\label{fig:MNIST:SI}
\end{figure*}


%

\end{document}